\documentclass[a4paper,fleqn]{cas-dc}

\usepackage[numbers]{natbib}

\usepackage{amsmath,amsfonts}
\usepackage{algorithmic}
\usepackage{array}
\usepackage[caption=false,font=normalsize,labelfont=sf,textfont=sf]{subfig}
\usepackage{textcomp}
\usepackage{stfloats}
\usepackage{url}
\usepackage{verbatim}
\usepackage{graphicx}

\usepackage{booktabs}
\usepackage{bm}
\usepackage{epsfig}
\usepackage{comment}
\usepackage{multirow}
\usepackage{makecell}
\usepackage[linesnumbered,ruled]{algorithm2e}
\newcommand{\rharpoon}{\overset{\rightharpoonup}}
\newcommand{\lharpoon}{\overset{\leftharpoonup}}

\def\tsc#1{\csdef{#1}{\textsc{\lowercase{#1}}\xspace}}
\tsc{WGM}
\tsc{QE}

\begin{document}
\let\WriteBookmarks\relax
\def\floatpagepagefraction{1}
\def\textpagefraction{.001}

\shorttitle{Differentiated Relevance Embedding for Group-based Referring Expression Comprehension}

\shortauthors{F. Chen}  

\title [mode = title]{Differentiated Relevance Embedding for Group-based Referring Expression Comprehension}

\author[1]{Fuhai~Chen}
\ead{chenfuhai3c@163.com}
\affiliation[1]{organization={Department of Computing, The Hong Kong Polytechnic University},
            city={HK},
            country={China}}

\author[2]{Xuri~Ge}
\cormark[1]
\ead{x.ge.2@research.gla.ac.uk}
\affiliation[2]{organization={School of Computing Science, University of Glasgow},
            country={UK}}

\author[3]{Xiaoshuai~Sun}
\ead{xssun@xmu.edu.cn}
\affiliation[3]{organization={Department of Artificial Intelligence, School of Informatics, Xiamen University},
            country={China}}

\author[4]{Yue~Gao}
\ead{gaoyue@tsinghua.edu.cn}
\affiliation[4]{organization={School of Software, Tsinghua University},
            country={China}}

\author[5]{Jianzhuang~Liu}
\ead{liu.jianzhuang@huawei.com}
\affiliation[5]{organization={Huawei Noahs Ark Lab},
            city={Shenzhen},
            country={China}}

\author[6]{Fufeng~Chen}
\cormark[2]
\ead{chenfufeng@fzidt.com}
\affiliation[6]{organization={Fuzhou Institute for Data Technology},
            country={China}}
            
\author[1]{Wenjie~Li}
\cormark[2]
\ead{cswjli@comp.polyu.edu.hk}

\cortext[1]{Co-first author with equal contribution}
\cortext[2]{Corresponding authors}

\begin{abstract}
Referring expression comprehension aims to locate a specific object in an image referred by a natural language expression, where the key lies in capturing the cross-modal relevance between the expressions and their corresponding objects. 
Existing works typically model the cross-modal relevance by estimating the attribute-oriented triplet ranking loss in each image, where the anchor object/expression and their positive expression/object have the same attribute as the negative expression/object, but with different attribute values, \emph{e.g.} \emph{red} and \emph{black} for the attribute \emph{clothing color}. 
These objects/expressions are exclusively utilized to learn the implicit representation of the attribute by a pair of different values,  
which however impedes the accuracies of the attribute representations, expression/object representations, and their cross-modal relevances since each anchor object/expression usually has multiple attributes while each attribute usually has multiple potential values. 
To effectively learn the cross-modal relevance, we investigate a novel REC problem named Group-based REC, where each object/expression is simultaneously employed to construct the multiple triplets among the semantically similar images. 
To tackle the explosion of the negatives and the differentiation of the anchor-negative relevance scores, 
we propose a novel multi-group self-paced relevance learning schema (termed MSRL) for the cross-modal alignment,
where the within-group object-expression pairs are adaptively assigned with different priorities based on their cross-modal relevance scores during model training. 
Since the average cross-modal relevance varies a lot across different groups, we further design an across-group relevance constraint to balance the bias of the group priority. 
Quantitative experiments on three widely-used REC benchmarks (RefCOCO, RefCOCO+, and RefCOCOg) demonstrate the superiority of the proposed MSRL method against the state-of-the-art methods. Extensive ablative experiments are conducted to further demonstrate the effectiveness of the adaptive learning strategy on the differentiated within-group and across-group relevance.
\end{abstract}


\begin{highlights}
\item Referring expression comprehension
\item Self-paced Relevance learning
\item Multiple image groups
\item Adaptive learning
\end{highlights}


\maketitle

\section{Introduction}

\begin{figure}
\centering
\epsfig{file=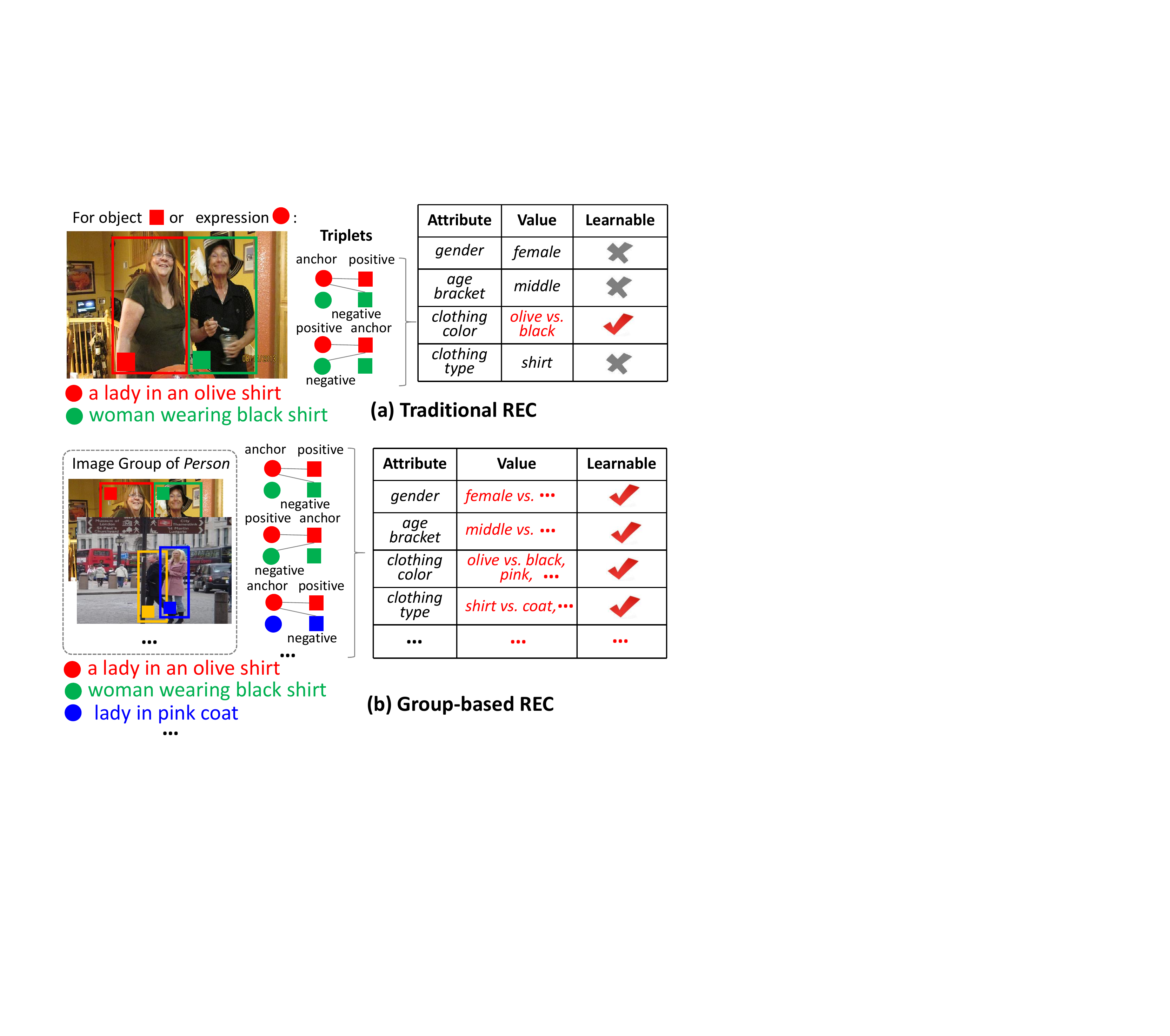, width=1.0\linewidth}
\caption{Two schemes of referring expression comprehension (REC). In traditional REC, the triplets with the matched (anchor-positive) and unmatched (anchor-negative) expression-object pairs are constructed from individual image towards the differences of attributes. In our Group-based REC, the negatives come from the images of the same group. There are more negatives to cover more learnable attributes with more complete differences of attribute values compared to the anchor.
There are two issues for Group-based REC: (1) the cross-modal relevance scores of the anchor-negative pairs are differentiated within each image group, and (2) the average cross-modal relevance scores of the matched/unmatched pairs are differentiated across different image group.\label{fig:differentiated diversity}}
\end{figure}

Referring expression comprehension (REC) aims to locate a specific object in an image referred by a given natural language expression, \emph{e.g.} locating \emph{a lady in an olive shirt} among all the persons in the first image of Fig. \ref{fig:differentiated diversity}. 
With the increasing demand of the ternary fusion intelligence among humans, machines, and the physical world, REC has attracted extensive research attention, serving as the foundation of many multi-modal AI problems such as visual dialog \cite{tu2021learning,chen2022utc}, robotic navigation \cite{zhao2022target,qiao2023hop+}, visual Q\&A \cite{gupta2022swapmix,guo2023sparse}, and video surveillance \cite{wan2020intelligent,du2022omg}. 

The challenge of REC lies in modeling the cross-modal relevance between the expressions and objects for the accurate cross-modal alignment, where each expression is expectedly more relevant to its corresponding object than other objects of the same category. 
Existing works \cite{mao2016generation,yu2016modeling,liu2017referring,hu2017modeling,yu2018mattnet,yang2019dynamic,liu2019improving, chen2020cops, cheng2021exploring, shen2022continual} mainly adopt the triplet-ranking-loss based alignment method to maximize the matching score of each matched  pair (anchor-positive) between a selected expression/object and its corresponding object/expression while minimizing the score of the unmatched pair (anchor-negative) between the selected expression/object and the rest object/expression in each image of REC dataset \cite{yu2016modeling,mao2016generation}. 
Each of these images contains an attribute-oriented anchor-positive-negative triplet, 
where the anchor, the positive, and the negative have the same attribute but the latter's attribute value is different from the first two. 
For example, the attribute \emph{clothing color} in the first image of Fig. \ref{fig:differentiated diversity} has different values, \emph{olive} and \emph{black} for the anchor/positive and the negative respectively. 
In this way, the attribute and its values are learned to encourage the anchor and the positive close while forcing the anchor and the negative apart. 
However, due to the inherent semantic richness of the objects and expressions, each object/expression usually has multiple attributes and each attribute usually has multiple values, \emph{e.g.}, the \emph{lady} in Fig. \ref{fig:differentiated diversity} has the attributes \emph{clothing color}, \emph{clothing type}, \emph{gender}, \emph{age bracket}, \emph{etc}, and the value of the attribute \emph{clothing color} could be \emph{olive}, \emph{black}, or \emph{red}, \emph{etc}. 
When the limited negatives face with the multiple attributes and multiple attribute values, it would be hard to fully learn the attribute and the expression/object, which suppresses the accuracies of their representations and their relevances for the cross-modal alignment as reflected in Fig. \ref{fig:differentiated diversity}.
For example, the value \emph{olive} of the attribute \emph{clothing color}, the factor of the anchor-positive relevance is supposed to be learned based on \emph{olive} \emph{vs.} others with more negatives rather than only \emph{olive} \emph{vs.} \emph{black}.
Additionally, the anchor and the positive are relevant actually due to not only the same \emph{clothing color} but also the same \emph{clothing type}, the same \emph{gender}, \emph{etc}, where more negatives with various attributes are expected for the cross-modal relevance learning.

To tackle the above problems, we redefine the REC problem as the Group-based REC problem, 
where the existing data is fully utilized to augment the negatives by using each object/expression repeatedly for multiple triplets with various attributes.
Specially, Group-based REC aims to learn the cross-modal relevance with the negatives from the semantically similar images with the same object category. We name these category-specific image set as a group for simplicity. 
To conduct Group-based REC on these expressions and objects, one intuitive idea is to annotate all attributes and their combinations for all these expressions and objects, enumerate all the possible values for each attribute in the dataset, and conduct the precise cross-modal alignment with all the expression-object pairs covering all the differences of attribute, attribute combination, and attribute value. 
However, such approach distinctly increases the complexity of model training, reduces the flexibility of model on the complex sentences and scenes, and restricts the generalization of model.
%
Another intuitive way is to mix such a large number of negatives in the same group for each anchor and randomly sample them for the cross-modal alignment without distinction.  
However, as revealed in \cite{schroff2015facenet,wu2017sampling,kalantidis2020hard}, the ranking loss based modeling with huge amounts of data is sensitive to the different hardness levels of different anchor-negatives according to their distances when sampling among the multiple negatives, which affects the final performance. This motivates us to treat these negatives adaptively with the distinction according to the estimated cross-modal relevance scores of the anchor-negatives. 
Inspired by \cite{bengio2009curriculum,jiang2014self,kaushal2018learning}, where self-paced learning (SPL) is proposed to estimate the hardness of samples while training the classification model progressively fed with the samples from easy to hard, it's natural for us to consider an adaptive sampling and progressive learning strategy on Group-based REC.

To conduct Group-based REC with adaptive learning strategy, the key issues lie in two-fold. On one hand, different from the sample hardness in SPL, the hardness in Group-based REC derives from the cross-modal relevance between the anchor expression/object and the negative object/expression within each image or across images instead of the estimated sample-wise loss. Meanwhile, such hardness will affect the inportance of the relevance differences between the anchor-positive pairs and the anchor-negative pairs instead of the estimated sample-wise losses.
On the other hand, the average value of the cross-modal relevance varies a lot across different image groups. For example, in Fig. \ref{fig:differentiated diversity}, the average relevance value of the \emph{car} group is apparently larger than that of the \emph{person} group due to the subtle and non-rigid variations of the latter one. This is also reflected by Coefficient of Variation\footnote{Coefficient of Variation (CV) \cite{abdi2010coefficient} represents the ratio of the standard deviation to the mean, which is a statistical measure of the dispersion of data points in a data series around the mean.} 15.64\% over different groups, calculated based on the results of the representative REC method \cite{yu2018mattnet} on the standard dataset, RefCOCOg \cite{mao2016generation}. 
Such diversity of average relevance would probably lead the model to overfitting to certain groups if simply using the anchor-negative pairs based on their differentiated cross-modal relevance values without considering the group distinction.

Driven by the above insights, we propose a novel self-paced relevance learning method (termed MSRL) for Group-based REC. 
Our main innovations lie in two aspects. Firstly, inspired by the idea of self-paced learning \cite{kumar2010self,jiang2014easy,ren2017robust} where the model is learned progressively by using samples from easy to hard according to their sample-wise losses, 
we introduce an adaptive relevance learning strategy to progressively learn the differentiated within-group cross-modal relevance.
In such strategy, the within-group unmatched anchor-negative pairs are automatically assigned with different priorities based on the updated cross-modal relevance scores. 
%
Secondly, to relieve the priority bias derived from the inherent relevance variance of different groups,
we integrate an across-group cross-modal relevance constraint over the anchor-negative pairs of different groups into the adaptive REC model training to enforce a balanced training, where the across-group variety of the unmatched pairs is formulated via a relevance-based regularization term. 
We conduct quantitative experiments on three widely-used REC benchmarks (RefCOCO, RefCOCO+, and RefCOCOg), which demonstrate the superiority of the proposed MSRL method against the state-of-the-art methods and the effectiveness of the adaptive relevance learning strategy.

In summary, the paper makes the following contributions: (1) We investigate a new problem of REC, \emph{i.e.} Group-based REC, where the expressions/objects are correlated across different images. (2) We are the first to explore and utilize the difference of the cross-modal relevance scores of the object-expression pairs for the ranking-loss based REC training. (3) A novel adaptive progressive learning strategy, \emph{i.e.} multi-group self-paced relevance learning strategy (MSRL) is proposed to model the within-group and across-group cross-modal relevances for Group-based REC. (4) The proposed MSRL method achieves the state-of-the-art in three commonly-used REC benchmarks.

The rest of paper is organized as follows: Section \ref{sec:Related_Work} reviews the related work. In Section \ref{sec:GroupREC}, we introduce the proposed multi-group self-paced relevance learning method (MSRL) for Group-based REC. Experimental evaluations and analysis are given in Section \ref{sec:experiments}. And finally, we conclude this paper in Section \ref{sec:conclusion}.

\section{Related Work\label{sec:Related_Work}}

\subsection{Referring Expression Comprehension} 

Referring expression comprehension relies on the cross-modal relevance to localize the special object referred by a given expression in an image. To this end, current works mainly follow two mainstreams, \emph{i.e.} implicit and explicit cross-modal relevance modeling schemes, where the visual and the textual semantic are respectively represented indirectly and directly.
On one hand, some works follow the implicit cross-modal relevance modeling by the graph-based methods \cite{wang2019neighbourhood,yang2019dynamic,yang2019cross,yang2020graph,liu2020learning,chen2022understanding,wang2023referring}, the parser-based methods \cite{cirik2018using,liu2019learning,yang2020propagating}, and the transformer-based methods \cite{lu2019vilbert,su2020vl,lu202012,li2021referring,deng2021transvg,du2022visual,sun2022proposal}, where the semantic representation is learned associating with other objects under the guidance of the semantic knowledge derived from the text. For example, Yang \emph{et al.} \cite{yang2019cross} construct a language-guided visual relation graph with cross-modal attention and propagates the fused cross-modal information to obtain semantic contexts, where each object representation is learned based on the other objects and textual knowledge.
Cirik \emph{et al.} \cite{cirik2018using} use the external syntactic analysis to parse the input reference expressions and guide the model to learn the semantic representations of objects and their interpretable relationships under the syntactic structure. 
Lu \emph{et al.} \cite{lu2019vilbert} propose the representative transformer-based cross-modal representation learning method, \emph{i.e.} ViLBERT, where the object's and the expression's high-level representations are learned based on the intra-modal correlation and the inter-modal correlation via the multi-layer transformer encoder.

On the other hand, most of works proceed the explicit cross-modal relevance modeling by either the joint embedding methods \cite{mao2016generation,liu2017referring,yu2017joint,yu2018mattnet, liu2019improving, yang2020improving, chen2021ref,li2021bottom,sun2021iterative,suo2022rethinking} or the modular-based methods \cite{hu2017modeling, yu2018mattnet, chen2020cops, sun2021discriminative, shen2022continual}. For example, Mao \emph{et al.} \cite{mao2016generation} introduce the first CNN-LSTM approach for referring expression comprehension. MattNet \cite{yu2018mattnet}, the representative and fundamental method, combines joint embedding and the modular composition by designing a Faster-RCNN based visual and a Bi-LSTM based textual modular attention networks to extract the semantic features of the subjects, locations, and relationships in the images and the expressions. And it has so far proved competitive and widely served as the backbone due to the explicit representation of the rich semantic \cite{liu2019improving, chen2020cops, cheng2021exploring, shen2022continual}. 
However, due to the pre-trained model, the prior knowledge, and the construction of more within-modal and cross-modal relationships, the implicit cross-modal relevance modeling tends to achieve better REC performance than the explicit one. Despite the fact, the explicit modeling still attracts the research interest due to the independence and the inherency of the object representation without the distraction of the scene-specific context, which can be independently taken as the foundational representation for many related tasks, such as continue learning \cite{shen2022continual}, weakly supervised learning \cite{sun2021discriminative}, and logical reasoning \cite{cheng2021exploring}.
Driven by this, we conduct group-based REC based on the explicit cross-modal relevance modeling, where the negative objects and expressions can be treat equally regardless of whether they come from the same image as the anchor and rely on correlation with the anchor or not. 
To this end, we choose MattNet \cite{yu2018mattnet}, the representative method as the backbone of the explicit cross-modal relevance modeling, which covers both the joint embedding and the modular construction.

\subsection{Self-Paced Learning} 


Different from the traditional machine learning algorithms, where all the training examples are randomly fed to the model, curriculum learning (CL) focuses on the various complexities of data samples and the learning status of the current model, which forces to learn based on the samples from easy to hard, imitating human curricula. CL is first proposed by Bengio \emph{et al.} \cite{bengio2009curriculum} and developed by \cite{spitkovsky2010baby,pentina2015curriculum,shi2015recurrent,gong2016multi,graves2017automated} to measure the the hardness of sample based on the prior knowledge and schedule the time and the numbers of the harder samples to feed the \emph{learner}.
Inspired by human education \cite{tullis2011effectiveness}, self-paced learning, the main branch and the advanced alternative of curriculum learning, is designed to allow the \emph{learner} to plan and control its learning curriculum. SPL is first proposed by Kumar \emph{et al.} \cite{kumar2010self} and then developed on many practical problems \cite{tang2012self,jiang2014easy,zhang2015self,zhang2017spftn,zhou2018deep,zhang2019leveraging,zheng2020unsupervised,wei2021meta,klink2021probabilistic,gu2022balanced,alavi2022bi,yu2022deep}, where the models are expected to adaptively estimate the example-wise training loss as the criterion to select the training samples from easy to hard. 
For example, 
Jiang \emph{et al.} \cite{jiang2014self} and Zhang \emph{et al.} \cite{zhang2017co} make the efforts to additionally integrate sample diversity in SPL for event/action recognition and saliency detection, respectively. The sample diversity is estimated according to the diversity of the sample loss, which also depends on the sample easiness.
Zhang \emph{et al.} \cite{zhang2019leveraging} proposal a collaborative self-paced learning method to select the samples with both the high instance-level confidence and the high image-level confidence for the model training of the weakly-supervised object detection. 
Yu \emph{et al.} \cite{yu2022deep} propose a self-paced refinement method to gradually drop the suspicious samples and focus on the desired samples to train the video anomaly detection model.
Most of the traditional works conduct sampling based on the measurement of the sample hardness in SPL, 
which take the loss of the sample in each training iteration as the estimation of its hardness. Although Jiang \emph{et al.} \cite{jiang2014self} and Zhang \emph{et al.} \cite{zhang2017co} make the efforts to additionally integrate sample diversity in SPL, the sample diversity is estimated according to the diversity of the sample loss, which also depends on the sample hardness. However, these traditional loss-guided SPL methods are infeasible in Group-based REC because: 
(1) The measure of hardness in Group-based REC orientates the expression-object pairs across/within images instead of the images or other specific instances in the dataset.
(2) Group-based REC focuses on the cross-modal relevance of the anchor-negative pair across/within images instead of the estimation loss between the predicted label and the ground-truth label of each instance.
(3) Group-based REC measures the anchor-negative relevance and adaptively schedules the learning on the relevance difference between the anchor-positive pair and the anchor-negative pair. This is different from SPL which estimating the loss and schedules the learning on the estimated loss, since the triplet ranking loss is unstable compared to cross-entropy loss due to the sampling of triplet items as revealed in \cite{hermans2017defense}. 
(4) Different from SPL, where the measure of hardness is restricted to the individual sample each time, Group-based REC involves the multiple anchors in different groups, where the inherent relevance variances among groups need to be considered for the hardness measuring. 

\begin{figure*}
\centering
\epsfig{file=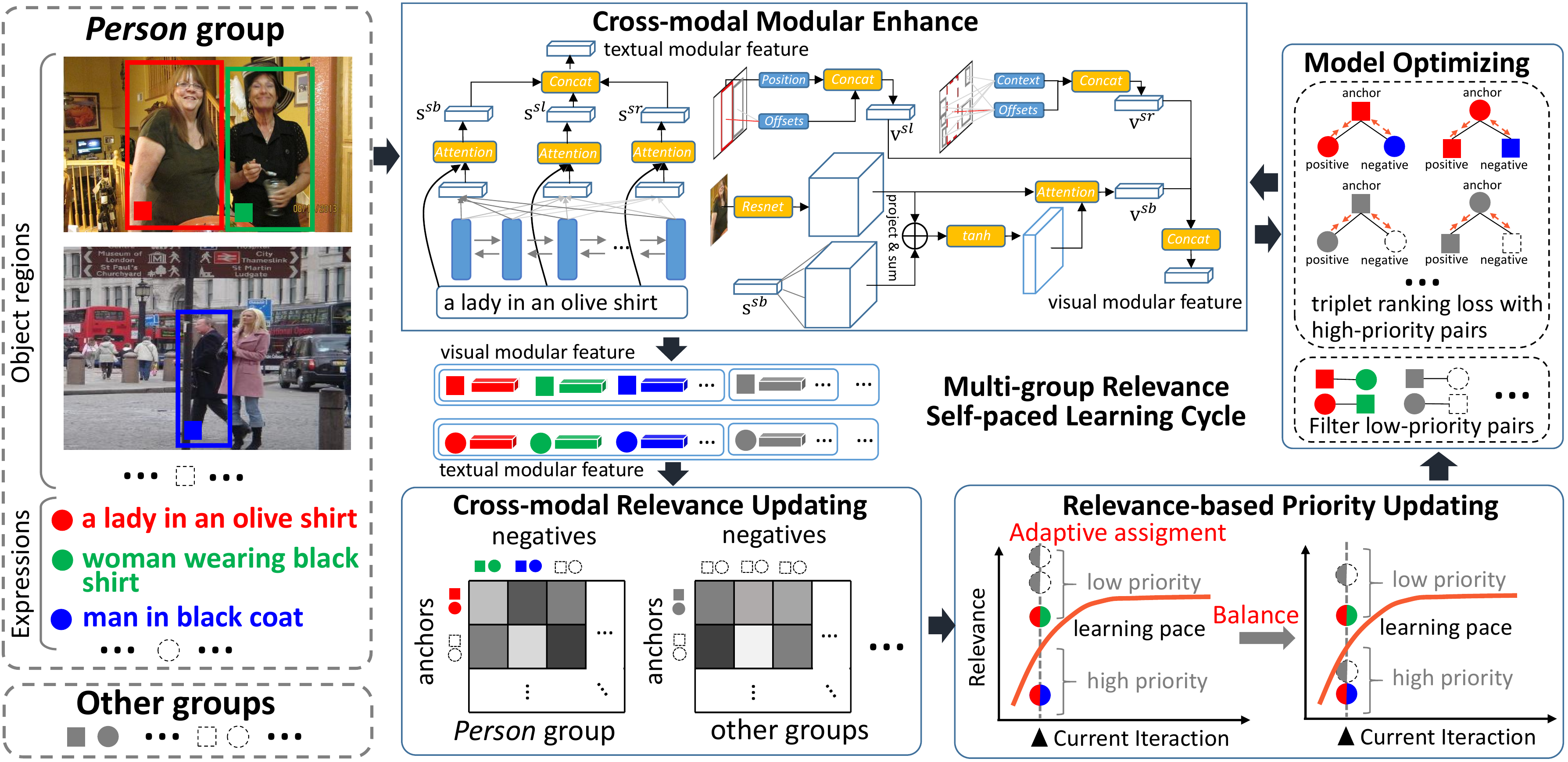, width=1.0\linewidth}
\caption{The framework of multi-group self-paced relevance learning schema (MSRL) for Group-based REC. 
MSRL first takes the object regions and their expressions from multiple image groups as the inputs during each training iteration. 
The visual and the textual modular features are then extracted in each iteration via the cross-modal modular enhance module, where the subject is enhanced with the spatial and semantic contexts to strengthen the visual and textual specificity. 
The cross-modal relevance scores are estimated between the anchors and the negatives and used to adaptively assign the anchor-negative pairs the training priorities within each image group. Simultaneously, MSRL balances the relevance bias across different groups via a relevance based constraint.
Finally, a priority based triplet ranking loss is minimized to optimize the parameters of cross-modal modular enhance module. The above steps are proceeded circularly as the model is fed by the anchor-negative pairs from low-relevance to high-relevance.
\label{fig:framework}}
\end{figure*}

\section{Method\label{sec:GroupREC}}

This paper proposes multi-group self-paced relevance learning method (MSRL) for Group-based REC.
In this section, we first describe the problem formulations of REC and the proposed Group-based REC in Section \ref{subsec:problem_formulation}. 
Next, we introduce the adaptive priority assignment of MSRL in Section \ref{subsec:relevance_updating}, where the feature representation, relevance estimation and priority updating are introduced. 
Finally, we formulate MSRL objective and describe its optimizing strategy in Section \ref{subsec:MSRL_objective}, where the self-paced relevance learning process and the across-group relevance constraint are introduced.

\subsection{Problem Formulation\label{subsec:problem_formulation}}

\textbf{Traditional REC}. As the foundation of Group-based REC, REC aims to locate a particular object in an image referred by a given natural language expression, which is typically set as a ranking-based retrieval problem: given an image, a query expression (sentence), and a set of object regions extracted from the image, the proposal with the highest matching score is chosen as the target object according to the matching score between each object region and the query. During model training, each triplet with an anchor, a positive and a negative is constructed to feed the ranking model to closen the anchor and positive while forcing the anchor and the negative apart. 

Suppose there are $N$ matched (anchor-positive) region-expression pairs, the triplet ranking loss can be formulated based on the object regions $\{v_i|i= 1,\ldots,N\}$ and the expressions $\{s_i|i= 1,\ldots,N\}$:

\vspace{-3mm}
\begin{align}
\label{eq:original_loss_function}
\min_{\bm{\theta}} & \sum_{i=1}^{N}\big[\max\big(0,\Delta+{F}(v_i,s_j;\bm{\theta})-{F}(v_i,s_i;\bm{\theta})\big) \nonumber \\
& + \max\big(0,\Delta+{F}(v_k,s_i;\bm{\theta})-{F}(v_i,s_i;\bm{\theta})\big)\big], \\
& s.t.\ i\neq j\ and\ i\neq k, \nonumber
\end{align}
\noindent
where $v_i$, $s_i$, and $s_j$ in the first $\max()$ denote the anchor object, the positive expression, and the negative expression, respectively. $s_i$, $v_i$, and $v_k$ in the second $\max()$ denote the anchor expression, the positive object, and the negative object, respectively.
$s_j$ and $v_k$ are randomly sampled from the same image as the anchor. $\Delta$ denotes the margin of the ranking loss. ${F}({v},{s};\bm{\theta})$ returns a matching score of $v$ and $s$ where their semantic features are learned and fed to a multi-layer perception layer (MLP) and an L2 normalization layer typically as set in \cite{yu2018mattnet}. $\bm{\theta}$ is the parameter set of the feature representation and the matching operation. Under such a max-margin pattern, ${F}$ outputs a high matching score for the matched (anchor-positive) region-expression pair $<v_i, s_i>$, while it outputs low matching scores for the unmatched (anchor-negative) region-expression pairs $<v_i, s_j>$ and $<v_k, s_i>$.
Note that the anchor, the positive, and the negative in each triplet derive from the same image and reflect the consistency and the difference on one attribute, which greatly limits the using scopes of the expressions and objects.

\textbf{Group-based REC}. The goal of Group-based REC is to learn a REC model upon the grouped region-expression pairs to locate the correct object referred by a given expression. In Group-based REC, we split the object regions $\{v_i|i= 1,\ldots,N\}$ and the expressions $\{s_i|i= 1,\ldots,N\}$ into $N_G$ groups, \emph{i.e.} $\{\mathcal{G}_g|g=1,\ldots,N_G\}$ according to the subjectival categories of the expressions, \emph{e.g.} the \emph{person} group with \emph{person} category. 
These groups are named \textbf{image groups} since each image is only annotated with the expressions that belong to a certain subjectival category \cite{yu2016modeling,mao2016generation}.
To construct these image groups, we follow the list of the $N_G$ subject categories in the REC datasets, parse the expressions via Stanford Parser \cite{socher2011parsing} and locate the subject for the expressions and their object regions via pos-tag tool in NTLK \cite{loper2002nltk}.
Thus, we obtain the grouped regions and expressions in image groups, where the $g$-th group has $N_g$ pairs and is denoted as $\mathcal{G}_g = \{v_i^{(g)}, s_i^{(g)}| i=1, \ldots, N_g\}$.

To formulate the triplets, suppose there is a matched pair set $\mathcal{P}$ with $M$ randomly-sampled matched pairs in a certain batch, we re-index the pairs and formulate $\mathcal{P}$ as $\{<v_i^{(g)},s_i^{(g)}>|i=1,\ldots,M\ \text{and}\ v_i^{(g)},s_i^{(g)}\in \mathcal{G}_g\}_{g=1}^{N_G}$. 
Taking each region or expression in $\mathcal{P}$ as the anchor and its matched expression or region as the positive, we construct the unmatched pair set with $M'$ negatives from different groups by randomly sampling the negative expression $s_j$ or region $v_k$ from the same image group as the anchor.
The unmatched pair sets are formulated as $\mathcal{P}'_1=\{<v_i^{(g)},s_j^{(g)}>|v_i^{(g)} \in \mathcal{P}, s_j^{(g)}\in \text{Rand}(\mathcal{G}_g)\}_{g=1}^{N_G}$ and $\mathcal{P}'_2=\{<v_k^{(g)},s_i^{(g)}>|s_i^{(g)}\in \mathcal{P}, v_k^{(g)}\in \text{Rand}(\mathcal{G}_g)\}_{g=1}^{N_G}$, 
where we let $v_k^{(g)}$ (negative) different $v_i^{(g)}$ (anchor) and $s_j^{(g)}$ (negative) different from $s_i^{(g)}$ (anchor) for simplicity even though $k=i$ and $j=i$. 
To reduce the unmatched pairs with the inoperative negatives, we filter the pairs $<{v}_i,{s}_j>$ and $<{v}_j,{s}_i>$ according to the relative distance between $v_i$ and $v_j$: if $v_i$ and $v_j$ are in the same image and their intersection-over-union (IoU) is less than 0.5, then these two corresponding pairs will be filtered out, otherwise kept. 
This helps to capture the spatial difference in each unmatched pair and prevent to construct the unmatched pair with the anchor and the positive.

\begin{table}[t]
\centering
\caption{Notations and definitions}
\begin{tabular}{p{2.1cm}<{\raggedright}|p{6.0cm}<{\raggedright}}
\hline
Notation & Definition \\
\hline
\hline
$N$                 & the number of the matched region-expression pairs \\
$N_G$               & the number of image groups \\
$\mathcal{G}_g$     & the set of the matched region-expression pairs in the $g$-th group \\
$M$                 & the number of the anchors in a batch \\
$M'$                & the number of the negatives in a batch \\
$s$                 & the expression \\
$v$                 & the object region \\
$<v,s>$             & a region-expression pair \\
${\rm\mathbf{s}}\in R^{3d}$                    & the expression feature \\
${\rm\mathbf{v}}\in R^{3d}$                    & the object region feature \\
$\mathcal{P}$       & the matched pair set in a batch\\
$\mathcal{P}'_1$, $\mathcal{P}'_2$  & two unmatched pair sets for the negative expression and negative object region respectively in a batch\\
${\rm\mathbf{R}}^{(g)} \in R^{M\times M'}$   & the relevance matrix of the $g$-th group\\
${\rm\mathbf{U}}^{(g)}\in R^{M\times M'}$    & a binary priority matrix of the $g$-th group \\
$\bm{\mathcal{U}}=\{{\rm\mathbf{U}}^{(g)}\}_{g=1}^{N_G}$ & the set of the binary priority matrices over different groups \\
$\bm{\theta}$       & the model parameter set \\
$\Delta$            & the margin in the triplet ranking loss \\
$\alpha\in\{0,1\}$  & a switch coefficient to choose the visual or textual features for the relevance estimation \\
$\lambda_1$, $\lambda_2$        & the adaptive parameters to control the learning pace for the visual and textual relevances, respectively \\
$\gamma$            & the dynamically-changing coefficient of the across-group relevance constraint \\
$\tau$              & the trade-off coefficient of relevance threshold \\
$\mu_1$, $\mu_2$    & the updating strides of $\lambda_1$ and $\lambda_2$, respectively \\
$\eta$              & the rising rate of $\gamma$ \\
\hline
\end{tabular}
\end{table}

\subsection{Adaptive Priority Assignment\label{subsec:relevance_updating}}

In the image groups, the cross-modal relevances are estimated/updated to assign the priorities to the unmatched pairs in different training iteration, which involves three key issues: (1) constructing the groups and pairs in advance, (2) extracting the textual and visual semantic features, and (3) estimating the cross-modal relevance and assigning the priority to the pair. We itemize them as follows:

\noindent\textbf{Cross-modal Modular Enhance}. The expressions and the object regions mainly contain the information of subjects. Therefore, it is necessary to extract the features of the subject from each expression and each object region, respectively. Since the expressions and the object regions in the same image group mainly reflect the same subject, we further extract the spatial locations of the subjects and the semantic relationship between the subjects and the object contexts to strengthen the specificity of the expressions and the object regions.

Suppose there are an expression $s$ and a object region  $v$, we obtain their features via cross-modal attention and modular enhancing. 
For the textual modality, given an expression $s=\{w_t\}_{t=1}^{T}$ with $T$ words, we use a bi-directional LSTM (Bi-LSTM) \cite{graves2005framewise} to encode the context for each word. In particular, word $w_t$ is first embedded into a vector ${\rm\mathbf{e}}_t \in R^d$ via word embedding. Then the Bi-LSTM is applied to encode the whole expression upon $\{\rm\mathbf{e}_t\}_{t=1}^{T}$ as:

\vspace{-3mm}
\begin{gather}
\rharpoon{\rm\mathbf{h}}_t, \lharpoon{\rm\mathbf{h}}_t = \text{Bi-LSTM}({\rm\mathbf{e}}_t,\rharpoon{\rm\mathbf{h}}_{t-1}, \lharpoon{\rm\mathbf{h}}_{t+1}), \\
{\rm\mathbf{h}}_t = {\rm\mathbf{W}_h}[\rharpoon{\rm\mathbf{h}}_t,\lharpoon{\rm\mathbf{h}}_t],
\end{gather}
\noindent
where ${\rm\mathbf{W}_h}\in R^{d\times 2d}$ is a weight matrix to combine the bi-directional features. ${\rm\mathbf{h}}_t \in R^{d}$ is the $t$-th hidden feature derived from two directional hidden features $\rharpoon{\mathbf{h}}_t$ and $\lharpoon{\mathbf{h}}_t$. $[\cdot,\cdot]$ is a concatenation operation. Finally, we compute the attention weights $\{a_{t}^{m}|t=1, \ldots, T, and \ m\in\{sb,sl,sr\}\}$ of the words for the subject (sb), the spatial location (sl), and the semantic relationship (sr). With the attention weights, the weighted sum of word embedding is made to produce the modular features ${\mathbf{s}}^{sb}$, ${\mathbf{s}}^{sl}$, ${\mathbf{s}}^{sr} \in R^{d}$. We formulate this process as:

\vspace{-3mm}
\begin{equation}
a_{t}^{m} = \frac{exp({\rm\mathbf{w}}_m^{\mathsf{T}}{\rm\mathbf{h}}_t)}{\sum_{k=1}^Texp({\rm\mathbf{w}}_m^{\mathsf{T}}{\rm\mathbf{h}}_k)},\  {\rm\mathbf{s}}^{m} = \sum_{t=1}^Ta_{t}^{m}{\rm\mathbf{e}}_t,
\end{equation}
\noindent
where ${\rm\mathbf{w}}_m \in R^{d}$ denotes the transform vector for a certain modular. Subsequently, the modular features are concatenated into a feature vector ${\mathbf{s}}=[{\rm\mathbf{s}}^{sb},{\rm\mathbf{s}}^{sl},{\rm\mathbf{s}}^{sr}]$ as the feature of the expression $s$.

For the visual modality, we first extract the subject feature for each object region and then produce the features of the spatial location and the semantic relationship based on the subject. For the subject, we first extract the feature maps ${\mathbf{V}}\in R^{c\times B}$ from the forth convolution layers of ResNet \cite{he2016deep} with grid size $B=7\times7$ and channel $c$. Then, we compute the attention feature ${\mathbf{H}}_a\in R^{d\times B}$ and attention weight ${\rm\mathbf{A}} \in R^B$ of subject on the grid:

\vspace{-5mm}
\begin{gather}
{\rm\mathbf{H}}_a = tanh({\rm\mathbf{W}}_v{\rm\mathbf{V}} + {\rm\mathbf{W}}_s{\rm\mathbf{S}}^{sb}),\\
{\rm\mathbf{A}} = softmax({\rm\mathbf{w}}_a^{\mathsf{T}}{\rm\mathbf{H}}_a),
\end{gather}
\noindent
where ${\mathbf{S}}^{sb}\in R^{d\times B}$ is obtained by expanding ${\mathbf{s}}^{sb}$ over $B$. ${\mathbf{W}}_v\in R^{d\times c}$, ${\rm\mathbf{W}}_s\in R^{d\times d}$, and ${\mathbf{w}}_a\in R^{d}$ are the parameters of the non-linear mapping. Finally, the visual subject feature is computed by the weighted sum of $\mathbf{V}$ over $B$, \emph{i.e.}, ${\mathbf{v}}^{sb} = \sum_{i=1}^B{\mathbf{A}}_i{\mathbf{V}}_{\cdot,i}$ with shape $R^d$.

For the spatial location, we use a 5-dim vector ${\rm\mathbf{l}}$ to encode the top-left position $(x_{1},y_{1})$, the bottom-right position $(x_{2},y_{2})$, and the region area $w\times h$ relative to the image size $W\times H$, \emph{i.e.} ${\mathbf{l}}=[\frac{x_{1}}{W},\frac{y_{1}}{H},\frac{x_{2}}{W},\frac{y_{2}}{H},\frac{w\cdot h}{W\cdot H}]$. Additionally, we compute the relative location between the candidate region and each of its five surrounding regions (at most and with the same category) via their position offsets and area ratio, \emph{i.e.} $\delta {\mathbf{l}}_{j}=[\frac{[\Delta x_{1}]_{j}}{w},\frac{[\Delta y_{1}]_{j}}{h},\frac{[\Delta x_{2}]_{j}}{w},\frac{[\Delta y_{2}]_{j}}{h},\frac{w_j\cdot h_j}{w\cdot h}]$ (the $j$-th surrounding region). $\{\delta {\mathbf{l}}_{j}\}_{j=1}^5$ is then concatenated into $\delta 
 {\mathbf{l}}$. We finally compute the visual feature on spatial location, \emph{i.e.} ${\mathbf{v}}^{sl} = {\mathbf{W}}_l[{\mathbf{l}};\delta {\mathbf{l}}] + {\mathbf{b}}_l$ with the transform parameters ${\mathbf{W}}_l \in R^{d\times 10}$ and ${\mathbf{b}}_l \in R^{d}$.

For the semantic relationship, we first select five arbitrary surrounding regions $\{v_j\}_{j=1}^5$ of the current candidate region, where their features of the final fully-connected layer are taken as the context features $\{{\mathbf{v}}_{j}^{cxt}\}_{j=1}^5$. We then compute the corresponding position offsets of the each surrounding region relative to the current region is computed by $\delta {\rm\mathbf{l}}_{j}'=[\frac{[\Delta x_{1}]_{j}}{w},\frac{[\Delta y_{1}]_{j}}{h},\frac{[\Delta x_{2}]_{j}}{w},\frac{[\Delta y_{2}]_{j}}{h},\frac{w_j\cdot h_j}{w\cdot h}]$. The final semantic relationship feature ${\mathbf{v}}^{sr}$ is computed by linearly mapping ${{\mathbf{v}}}_{j}^{sr} = {\mathbf{W}}_r[{\mathbf{v}}_{j}^{cxt};\delta {\mathbf{l}}_{j}'] + {\mathbf{b}}_r$ and concatenating $\{{{\mathbf{v}}}_{j}^{sr}\}_{j=1}^5$. The above three modular features are concatenated into a final visual feature, \emph{i.e.} ${\mathbf{v}} = [{\mathbf{v}}^{sb},{\mathbf{v}}^{sl},{\mathbf{v}}^{sr}]$.

\noindent\textbf{Cross-modal Relevance Estimation}. The relevance scores of the matched/unmatched region-expression pairs are estimated based on the above visual and textual features. 
Since the negative object regions sometime come from detection instead of annotation while only the expressions are given for some negatives in practice, thus the negative object region or its expression may be missing. Suppose there are the anchor expression $s_i^{(g)}$, the anchor object region $v_i^{(g)}$, the negative expression $s_j^{(g)}$, and the negative object region $v_k^{(g)}$ in two unmatched pairs $<v_i^{(g)},s_j^{(g)}>$ and $<v_k^{(g)},s_i^{(g)}>$ of the $g$-th image group, the corresponding negative region $v_j^{(g)}$ and the negative expression $s_k^{(g)}$ tend to be missing in the dataset. To deal with the modality missing of the negatives, we estimate the relevance score over different modalities robustly as below:

\vspace{-3mm}
\begin{equation}
\label{eq:similarity_computing}
\resizebox{.91\hsize}{!}{${\rm\mathbf{R}}_{i,j}^{(g)} = \alpha\frac{[{\rm\mathbf{v}}_i^{(g)}]^{\mathsf{T}}{\rm\mathbf{v}}_j^{(g)}}{\sum_{i', j'}^{M,M'}[{\rm\mathbf{v}}_{i'}^{(g)}]^{\mathsf{T}}{\rm\mathbf{v}}_{j'}^{(g)}} + (1-\alpha)\frac{[{\rm\mathbf{s}}_i^{(g)}]^{\mathsf{T}}{\rm\mathbf{s}}_j^{(g)}}{\sum_{i',j'}^{M,M'}[{\rm\mathbf{s}}_{i'}^{(g)}]^{\mathsf{T}}{\rm\mathbf{s}}_{j'}^{(g)}},$}
\end{equation}
\noindent
where ${\mathbf{R}}^{(g)} \in R^{M\times M'}$ is the relevance matrix of the sampled anchor-negative pairs in the the $g$-th image group $\mathcal{G}_g$. ${\mathbf{R}}_{i,j}^{(g)}$ denotes the relevance score between the $i$-th anchor region/expression and the $j$-th negative expression/region in the pair $<v_i^{(g)},s_j^{(g)}>$ or $<v_j^{(g)},s_i^{(g)}>$ of $\mathcal{G}_g$. Especially, if there are $v_i^{(g)}\not\in \mathcal{G}_g$, $v_j^{(g)}\not\in \mathcal{G}_g$, $s_i^{(g)}\not\in \mathcal{G}_g$, or $s_j^{(g)}\not\in \mathcal{G}_g$, we fill ${\rm\mathbf{R}}_{i,j}^{(g)}$ with $Inf$ and skip any subsequent operations about it for the executable priority assignment in $\mathcal{G}_g$. $\alpha\in\{0,1\}$ is a switch coefficient to choose the visual or textual features for the relevance estimation. 

\noindent\textbf{Relevance-based Priority Updating}. To assign the priorities to the anchor-negative pairs for training, we design an adaptive threshold to assign the binary value to the pair $<v_i^{(g)},r_j^{(g)}>$ according to its ${\rm\mathbf{R}}_{i,j}^{(g)}$:

\begin{equation}
{\rm\mathbf{U}}_{i,j}^{(g)} =
\begin{cases}
1, \alpha=1\ \text{and}\ {\rm\mathbf{R}}_{i,j}^{(g)} < \lambda_1 + \tau\gamma, \\
1, \alpha=0\ \text{and}\ {\rm\mathbf{R}}_{i,j}^{(g)} < \lambda_2 + \tau\gamma, \\
0, \text{otherwise},
\end{cases}
\label{eq:diversityweight_computing}
\end{equation}
\noindent
where ${\rm\mathbf{U}}^{(g)}\in R^{M\times M'}$ is a binary priority matrix corresponding to ${\mathbf{R}}^{(g)}$. The pair $<v_i^{(g)},r_j^{(g)}>$ is selected for training if ${\rm\mathbf{U}}_{i,j}^{(g)}=1$, otherwise discarded in the current iteration. 
%
$\lambda_1$ and $\lambda_2$ are the adaptive parameters to control the learning pace for $\alpha\in\{0,1\}$. As $\lambda_1$ and $\lambda_2$ grow, the number of the pairs with higher relevance scores is dynamically injected to the training list. That is, the low-relevance pairs are preferentially used in the early training stage while the higher-relevance pairs are gradually utilized subsequently. 
$\gamma$ is the dynamically-changing coefficient of the across-group relevance constraint in Eq. \ref{eq:final_loss_function} subsequently, which makes more various pairs to be utilized with the across-group relevance balancing. 
$\tau$ is a fixed trade-off coefficient. With the training iteration, $\lambda_1$, $\lambda_2$ and $\gamma$ are updated as below:

\vspace{-3mm}
\begin{gather}
\lambda_*\leftarrow \lambda_*+\frac{\mu_*}{M\cdot M'}\sum_{g=1}^{N_G}\sum_{i=1}^{M}\sum_{j=1}^{M'}\max(0,1-{\rm\mathbf{R}}_{i,j}^{(g)}),\label{eq:update_lambda} \\
\gamma\leftarrow \eta\gamma, \label{eq:update_gamma}
\end{gather}
\noindent
where $*$ denotes 1 or 2. $\lambda_*$ is updated according to the mean value of the counter-relevance matrices with the stride coefficient $\mu_*$. This allows the model to adjust the stride of the learning pace: (1) When the regions and the expressions of the selected pairs totally remain lowly relevant, the training will be easier due to the significant differences, and thus the model is able to lean from the pairs with more various relevance scores as reflected in Eq. \ref{eq:diversityweight_computing}. (2) With the relevance learning and the total relevance increasing, there exist more various relevances to digest and the model begins to slow the increasement of the learning stride down to approach the balance. For $\gamma$, it rises with the rate $\eta$ during training. To prevent $\lambda_*$ and $\gamma$ from rising out of control, we set their upper bounds with 1.

\subsection{MSRL Objective\label{subsec:MSRL_objective}}

\begin{algorithm}[t]
  \caption{Training MSRL for Group-based REC}
  \label{alg:algorithm_DMG-SPL}
  \KwIn{The matched pair sets in $\{\mathcal{G}_g\}_{g=1}^{N_G}$} 
  \KwOut{The parameter set $\bm{\theta}^*$}
  Initialize the parameters $\bm{\theta}$, $\lambda_1$, $\lambda_2$ and $\gamma$; \tcp{\footnotesize Details are given in Section \ref{subsec:settings}.}
  Initialize $\bm{\mathcal{U}}$ as Eq. \ref{eq:diversityweight_computing}\;
  \While{not converged}
  {
  \tcp{\footnotesize Pair Construction:}
  Construct $\mathcal{P}$ from $\{\mathcal{G}_g\}_{g=1}^{N_G}$ by randomly sampling\;
  Initialize $\mathcal{P}'_1$, $\mathcal{P}'_2$ = $\emptyset$\;
  \For{i := 1 \KwTo $M$}
  {
  $\mathcal{P}'_1=\mathcal{P}'_1\cup\{<v_i^{(g)},s_j^{(g)}>|v_i^{(g)}\in \mathcal{P}, s_j^{(g)}\in \text{Rand}(\mathcal{G}_g)\}_{g=1}^{N_G}$\;
  $\mathcal{P}'_2=\mathcal{P}'_2\cup\{<v_k^{(g)},s_i^{(g)}>|s_i^{(g)} \in \mathcal{P}, v_k^{(g)}\in \text{Rand}(\mathcal{G}_g)\}_{g=1}^{N_G}$\;
  }
  \tcp{\footnotesize Updating $\bm{\theta}$ and $\bm{\mathcal{U}}$:}
  Update $\bm{\theta}^*=\arg\min_{\bm{\theta}}\textit{E}(\bm{\mathcal{U}}^*,\bm{\theta};\lambda,\gamma,\mathcal{P},\mathcal{P}'_1,\mathcal{P}'_2)$ by propagating gradient\; 
  Update $\bm{\mathcal{U}}^*=\arg\min_{\bm{\mathcal{U}}}\textit{E}(\bm{\mathcal{U}},\bm{\theta}^*;\lambda,\gamma,\mathcal{P},\mathcal{P}'_1,\mathcal{P}'_2)$ as:\\
  (1) Compute $\{{\rm\mathbf{R}}^{(g)}\}_{g=1}^{N_G}$ as Eq. \ref{eq:similarity_computing}\; 
  (2) Update $\bm{\mathcal{U}}$ as Eq. \ref{eq:diversityweight_computing}\;
  \tcp{\footnotesize Updating The Adaptive Parameters:}
  Update $\lambda_1$ and $\lambda_2$ as Eq. \ref{eq:update_lambda}\;
  Update $\gamma$ as Eq. \ref{eq:update_gamma}\;
  }
\end{algorithm}

The proposed MSRL aims to learn the differentiated within-group and across-group cross-modal relevances in an adaptive learning schema for Group-based REC. The main ideas of MSRL lie in two aspects: (1) the model is trained by using lowly relevant unmatched pairs, prior to using highly relevant ones within each group, which can be regarded as learning a coarse-to-fine pattern step-by-step. 
(2) the relevances of the unmatched pairs from different groups should be balanced to relieve the bias of the group priority when their expression-object pairs are utilized in each training iteration. 
For the first idea, we design a specific learning schema, \emph{i.e.}, \emph{Self-paced Relevance Learning}, which uses the relevance updated in Section \ref{subsec:relevance_updating} to guided the self-paced learning. For the second ideas, we use a regularization term of the across-group relevance to reduce the group-wise relevance sparsity.

\noindent\textbf{Self-paced Relevance Learning}. In Group-based REC, the cross-modal relevance scores of the unmatched region-expression pairs are differentiated within each group, which leads to different effects of the pairs at different stages of model training. 
To learn the within-group cross-modal relevance, we update the relevance scores of the pairs based on their visual and textual features and assign the priorities to the pairs as aforementioned.
As a result, we obtain the priority matrices $\{{\mathbf{U}}^{(g)}\}_{g=1}^{N_G}$, which indicate whether the pairs are selected according to their relevance scores. We use $\bm{\mathcal{U}}$ to represent the set of these priority matrices over different image groups. Following this, we minimize a triplet ranking loss based on the pair priority in an adaptive manner:

\begin{equation}
\label{eq:final_loss_function}
\begin{aligned}
\textit{E} &= \!\!\!\!\! \sum_{\resizebox{.13\hsize}{!}{$p_{i,i}^{(g)},p_{i,j}^{(g)},p_{k,i}^{(g)}$}}\!\!\!\!\!\big[{\rm\mathbf{U}}_{i,j}^{(g)}Q(p_{i,i}^{(g)},p_{i,j}^{(g)};\bm{\theta}) + {\rm\mathbf{U}}_{i,k}^{(g)}Q(p_{i,i}^{(g)},p_{k,i}^{(g)};\bm{\theta})\big] \\
& -  \resizebox{.1\hsize}{!}{$\frac{\lambda_1 + \lambda_2}{2}$}\|\bm{\mathcal{U}}\|_{1}, s.t.\ p_{i,i}^{(g)}\in \mathcal{P},p_{i,j}^{(g)}\in \mathcal{P}'_1,p_{k,i}^{(g)}\in \mathcal{P}'_2,
\end{aligned}
\end{equation}
\noindent
where $p_{i,i}^{(g)}$, $p_{i,j}^{(g)}$, and $p_{k,i}^{(g)}$ denote the pairs $<v_i^{(g)},s_i^{(g)}>$, $<v_i^{(g)},s_j^{(g)}>$, and $<v_k^{(g)},s_i^{(g)}>$, respectively. $Q(p_{i,i}^{(g)},p_{i,j}^{(g)};\bm{\theta})$ and $Q(p_{i,i}^{(g)},p_{k,i}^{(g)};\bm{\theta})$ are two max-margin functions respectively formulated as below:

\vspace{-3mm}
\begin{equation}
\begin{cases}
\max\big(0,\Delta+{F}(v_i^{(g)},s_j^{(g)};\bm{\theta})-{F}(v_i^{(g)},s_i^{(g)};\bm{\theta})\big), \\
\max\big(0,\Delta+{F}(v_k^{(g)},s_i^{(g)};\bm{\theta})-{F}(v_i^{(g)},s_i^{(g)};\bm{\theta})\big), \\
\end{cases}
\end{equation}
\noindent
where $\bm{\theta}$ denotes the parameter set of the model that needs to be learned. $\Delta$ is the margin of the ranking loss. We guide the two-way sub-terms of the ranking loss in Eq. \ref{eq:final_loss_function} respectively with two binary priorities in ${\rm\mathbf{U}}^{(g)}$, 
which enables the model to selectively learn from different anchor-positive and anchor-negative pairs in different stages. 
${\rm\mathbf{U}}^{(g)}$ is controlled by the adaptive parameters $\lambda_1$ and $\lambda_2$ in Eq. \ref{eq:diversityweight_computing} and updated in each iteration via Eq. \ref{eq:update_lambda}. This allows that the anchor-negative pairs with high relevance scores are gradually injected into the ranking loss and replace the lower-relevance ones with $\lambda_*$ growing. 
In Eq. \ref{eq:final_loss_function}, $\frac{\lambda_1 + \lambda_2}{2}\|\bm{\mathcal{U}}\|_{1}$ is a regularizer. The $l_1$-norm is formulated as:

\begin{table*}[t]
\centering
\caption{Performance comparisons with the state-of-the-art methods on ground-truth regions. All values are in \%. The best results are marked in bold}
\begin{tabular}{p{3.3cm}<{\raggedright}|p{1.1cm}<{\centering}p{1.1cm}<{\centering}p{1.1cm}<{\centering}|p{1.1cm}<{\centering}p{1.1cm}<{\centering}p{1.1cm}<{\centering}|p{1.1cm}<{\centering}p{1.1cm}<{\centering}p{1.1cm}<{\centering}}
  \hline
  \multirow{2}{*}{Methods}  & \multicolumn{3}{c|}{RefCOCO} & \multicolumn{3}{c|}{RefCOCO+} & \multicolumn{3}{c}{RefCOCOg}\\
  \cline{2-10}
  & val & testA & testB & val & testA & testB & val* & val & test \\
  \hline
  MMI \cite{mao2016generation} & - & 63.15 & 64.21 & - & 48.73 & 42.13 & 62.14 & - & -\\
  NegBag \cite{nagaraja2016modeling} & 76.90 & 75.60 & 78.00 & - & - & - & - & - & 68.40  \\
  visdif+MMI \cite{yu2016modeling} & - & 73.98 & 76.59 & -  & 59.17 & 55.62 & 64.02 & - & - \\
  Luo \cite{luo2017comprehension} & - & 74.04 & 73.43 & - & 60.26 & 55.03 & 65.36 & - & - \\
  CMN \cite{hu2017modeling} & - & - & - & - & - & - & 69.30 & - & - \\
  Speaker/visdif \cite{yu2016modeling} & 76.18 & 74.39 & 77.30 & 58.94 & 61.29 & 56.24 & 59.40 & - & - \\
  S-L-R \cite{yu2017joint} & 79.56 & 78.95 & 80.22 & 62.26 & 64.60 & 59.62 & 72.63 & 71.65 & 71.92 \\
  Attr \cite{liu2017referring} & - & 78.05 & 78.07 & - & 61.47 & 57.22 & 69.83 & - & - \\
  VC \cite{zhang2018grounding} & - & 78.98 & 82.39 & - & 62.56 & 62.90 & 73.98 & - & - \\
  A-ATT \cite{deng2018visual} & 81.27 & 81.17 & 80.01 & 65.56 & 68.76 & 60.63 & 73.18 & - & - \\
  PLAN \cite{zhuang2018parallel} & 81.67 & 80.81 & 81.32 & 64.18 & 66.31 & 61.46 & 69.47 & - & - \\
  Multi-hop FiLM \cite{strub2018visual} & 84.9 & 87.4 & 83.1 & \textbf{73.8} & \textbf{78.7} & 65.8 & 71.5 & - & - \\
  MAttN \cite{yu2018mattnet} & 83.85 & 85.42 & 83.18 & 71.31 & 74.22 & 66.24 & - & 77.27 & 76.82 \\
  \hline
  MSRL-WG & 84.79 & 86.61 & 83.15 & 71.58 & 73.20 & 66.18 & - & 78.47 & 77.63 \\
  MSRL-AG & 84.63 & 85.88 & 82.95 & 71.44 & 73.15 & 66.25 & - & 78.37 & 77.43 \\
  MSRL & \textbf{85.92} & \textbf{87.50} & \textbf{83.67} & 72.10 & 73.95 & \textbf{66.47} & - & \textbf{79.15} & \textbf{78.42} \\
  \hline
\end{tabular}
\label{tab:comparison_sota_gt}
\end{table*}

\vspace{-3mm}
\begin{equation}
\|\bm{\mathcal{U}}\|_{1} = \sum_{g=1}^{N_G}\sum_{i=1}^{M}\sum_{j=1}^{M'}{\rm\mathbf{U}}_{i,j}^{(g)},\;
\end{equation}
\noindent
where the $l_1$-norm represents the number of the sampled pairs with the expected expression-object relevance levels as Eq. \ref{eq:diversityweight_computing}. 
With Eq. \ref{eq:final_loss_function} minimized, the regularizer is expected to be larger, where the cross-modal relevances of the selected unmatched pairs are forced to be low according to Eq. \ref{eq:diversityweight_computing}. With $\lambda_1$ and $\lambda_2$ growing, the model focuses more on the effect of regularizer since there are more high-relevance unmatched pairs.

\noindent\textbf{Objective with Across-group Relevance Constraint}. Since the relevance variance exists across different image groups, this leads to the priority bias during the priority assignment of pairs. To reduce the bias and to balance the model training, we regularize the priority matrix set $\bm{\mathcal{U}}$ in a $l_{F,1}$-norm term, formulated as:

\vspace{-3mm}
\begin{equation}
 \|\bm{\mathcal{U}}\|_{F,1} = \sum_{g=1}^{N_G}\|{\rm\mathbf{U}}^{(g)}\|_{F}, 
\end{equation}
\noindent
where $\|{\rm\mathbf{U}}^{(g)}\|_{F}$ is a Frobenius norm of ${\rm\mathbf{U}}^{(g)}$. As indicated in \cite{Ming2006Model}, the $l_{2,1}$-norm leads to a group-wise sparse representation of a vector if minimized. Inspired by this, we constrain $\bm{\mathcal{U}}$ in a $l_{F,1}$-norm and oppositely maximize this term to punish the group-wise sparsity, \emph{i.e.}, non-zero items of $\{{\rm\mathbf{U}}^{(g)}\}_{g=1}^{N_G}$ tend to be scattered across a large number of groups. That is, the across-group relevance will be balanced and the variety of group will be ensured in this regularization. Integrated with this across-group relevance regularization, the total objective of MSRL can be rewritten as below:

\vspace{-3mm}
\begin{equation}
\label{eq:final_final_loss_function}
\begin{aligned}
\textit{E} &= \!\!\!\!\! \sum_{\resizebox{.13\hsize}{!}{$p_{i,i}^{(g)},p_{i,j}^{(g)},p_{k,i}^{(g)}$}}\!\!\!\!\!\big[{\rm\mathbf{U}}_{i,j}^{(g)}Q(p_{i,i}^{(g)},p_{i,j}^{(g)};\bm{\theta}) + {\rm\mathbf{U}}_{i,k}^{(g)}Q(p_{i,i}^{(g)},p_{k,i}^{(g)};\bm{\theta})\big] \\
& -  \resizebox{.1\hsize}{!}{$\frac{\lambda_1 + \lambda_2}{2}$}\|\bm{\mathcal{U}}\|_{1} - \gamma\|\bm{\mathcal{U}}\|_{F,1}, \\
& s.t.\ p_{i,i}^{(g)}\in \mathcal{P},p_{i,j}^{(g)}\in \mathcal{P}'_1,p_{k,i}^{(g)}\in \mathcal{P}'_2,
\end{aligned}
\end{equation}
\noindent
where $\gamma$ is the dynamically-changing coefficient in the across-group relevance regularization. It is updated as Eq. \ref{eq:update_gamma}, which makes the model to increase the attention on the across-group relevance balancing. For model training, we minimize $\textit{E}(\bm{\mathcal{U}},\bm{\theta})$ in Eq. \ref{eq:final_final_loss_function} to optimize the parameter set $\bm{\theta}$ and the priority matrix set $\bm{\mathcal{U}}$ alternately.

\noindent\textbf{Optimization Strategy}. The objective function in the traditional SPL is convex while the one in the proposed MSRL is non-convex due to the $l_{F,1}$-norm of the priority matrix set $\bm{\mathcal{U}}$. Consequently, the traditional gradient-based methods cannot be directly applied to the optimization of $\bm{\mathcal{U}}^*=\arg\min_{\bm{\mathcal{U}}}\textit{E}(\bm{\mathcal{U}},\bm{\theta})$. To this end, we adopt an alternative convex search (ACS) algorithm \cite{kumar2010self,tang2012shifting} for the biconvex optimization of MSRL. In particular, the parameter set $\bm{\theta}$ and the priority matrix set $\bm{\mathcal{U}}$ are optimized in an alternative way as shown in Alg. \ref{alg:algorithm_DMG-SPL}. In each iteration, $\bm{\theta}$ is first optimized by the gradient-based method while keeping $\bm{\mathcal{U}}$ fixed. And $\bm{\mathcal{U}}$ is then updated via Eq. \ref{eq:diversityweight_computing} while keeping $\bm{\theta}$ fixed. The self-paced parameters $\lambda_1$, $\lambda_2$, and $\gamma$ are subsequently updated with strides $\mu_1$, $\mu_2$, and $\eta$ according to Eq. \ref{eq:update_lambda} and Eq. \ref{eq:update_gamma}, respectively.

\section{Experiments\label{sec:experiments}}

In this section, we first introduce the experimental settings, including pre-processing, parameter settings, datasets, evaluation protocols, and competitive methods. Next, we quantitatively compare the results of our proposed model to the state-of-the-art methods on REC. Finally, we conduct detailed model analysis on the within-group and the across-group relevances respectively.

\begin{table*}
\centering
\caption{Performance comparisons with the state-of-the-art methods on the fully-automatic comprehension task. All values are in \%. The first-place and the second-place results are marked in bold and with underline respectively}
\begin{tabular}{p{4.0cm}<{\raggedright}|p{1.2cm}<{\centering}p{1.2cm}<{\centering}p{1.2cm}<{\centering}|p{1.2cm}<{\centering}p{1.2cm}<{\centering}p{1.2cm}<{\centering}|p{1.2cm}<{\centering}p{1.2cm}<{\centering}}
  \hline
  \multirow{2}{*}{Methods} & \multicolumn{3}{c|}{RefCOCO} & \multicolumn{3}{c|}{RefCOCO+} & \multicolumn{2}{c}{RefCOCOg}\\
  \cline{2-9}
  & val & testA & testB & val & testA & testB  & val & test \\
  \hline
  Luo \cite{luo2017comprehension} & - & 67.94 & 55.18 & - & 57.05 & 43.33 &  - & - \\
  S-L-R \cite{yu2017joint} & 69.48 & 73.71 & 64.96 & 55.71 & 60.74 & 48.80 & 60.21 & 59.63 \\
  PLAN \cite{zhuang2018parallel} & - & 75.31 & 65.52 & - & 61.34 & 50.86 & - & - \\
  MAttN \cite{yu2018mattnet} & 75.82 & 80.75 & 69.58 & 65.41 & 70.35 & 56.12 & 66.85 & 66.66 \\
  DGA \cite{yang2019dynamic} & - & 78.42 & 65.53 & - & 69.07 & 51.99 & - & 63.28 \\
  RCCF \cite{liao2020real} & - & 81.06 & 71.85 & - & 70.35 & 56.32 & - & 65.73 \\
  One-Stage-BERT \cite{yang2019fast} & 72.05 & 74.81 & 67.59 & 55.72 & 60.37 & 48.54 & 59.03 & 58.70 \\
  One-stage-Base \cite{yang2020improving} & 76.59 & 78.22 & 73.25 & 63.23 & 66.64 & 55.53 &  64.87 & 64.87 \\
  One-stage-Large \cite{yang2020improving} & 77.63 & 80.45 & 72.30 & 63.59 & 68.36 & 56.81 & 67.30 & 67.20 \\
  IS-REC \cite{sun2021iterative} & - & 74.27 & 68.10 & - & 71.05 & \textbf{58.25} & - & \textbf{70.05} \\
  PFOS \cite{sun2022proposal} & \underline{78.44} & \underline{81.94} & \underline{73.61} & 65.86 & \textbf{72.43} & 55.26 & \underline{67.89} & 67.63 \\
  VGTR \cite{du2022visual} & \textbf{79.30} & \textbf{82.16} & \textbf{74.38} & 64.40 & 70.85 & 55.84 & 66.83 & 67.28 \\
  \hline
  MSRL-WG & 76.44 & 80.23 & 69.51 & 65.67 & 70.78 & 56.29 & 66.97 & 66.47 \\
  MSRL-AG & 76.83 & 80.50 & 69.64 & \underline{65.89} & 70.45 & 56.50 & 67.12 & 67.03 \\
  MSRL & 77.60 & 81.08 & 70.12 & \textbf{66.22} & \underline{71.14} & \underline{57.86} & \textbf{67.99} & \underline{67.67} \\
  \hline
\end{tabular}
\label{tab:comparison_sota_det}
\end{table*}

\subsection{Experimental Settings\label{subsec:settings}}

\noindent\textbf{Training Settings}. In this part, we introduce training settings on four aspects, including sample sizes, parameter settings, variable dimensions, and training details. For sample sizes, we set the anchor number $M$ and the negative number $M'$ in each batch as 10 and 60, where each anchor is paired with 6 negatives in $\{\mathcal{P}'_1, \mathcal{P}'_2\}$. For parameter settings, $\lambda_1$, $\lambda_2$, and $\gamma$ are all initialized to 0.5 and updated once every 1,000 iterations. $\mu_1$, $\mu_2$, $\tau$, and $\Delta$ are empirically set as 0.1. $\eta$ is set as 1.1. The weights in $\bm{\theta}$ are all initialized randomly in $[0,1)$. For variable dimensions, the word embedding size and hidden state size of the LSTM are set to 512. The outputs of all MLPs and FCs in the model are also set to be 512-dimensional. For training details, we adopt Adam \cite{kingma2014adam} as the gradient-based method, where the initial learning rate is set as $4e^{-4}$ and halved every 8,000 iterations after the first 8,000-iteration warm-up. To avoid overfitting, we regularize the word-embedding and output layers of the LSTM in the textual attention network using dropout with ratio of 0.5. We extract the object regions by Faster-RCNN \cite{ren2015faster}. The method is implemented in PyTorch, which takes roughly 57 training hours on NVIDIA GTX 1080 Ti GPU.

\noindent\textbf{Datasets}. We evaluate our method on three REC benchmark datasets, \emph{i.e.}, RefCOCO \cite{yu2016modeling}, RefCOCO+ \cite{yu2016modeling}, and RefCOCOg \cite{mao2016generation}, all of which are derived from the MSCOCO dataset \cite{lin2014microsoft}.

\noindent
(1) \emph{RefCOCO (UNC RefExp)} contains 142,209 referring expressions for 50,000 objects in 19,994 images from COCO. Since \emph{Person} is much more frequent than other objects in the dataset, the split is person vs. objects: images containing multiple persons are in ``Test A'' and images containing multiple objects from other categories are in ``Test B''.

\noindent
(2)  
\emph{RefCOCO+} has 141,564 expressions for 49,856 objects in 19,992 images from COCO, which is then extended using ReferitGame \cite{kazemzadeh2014referitgame}. 
Different from RefCOCO dataset, players (annotators) are not allowed to use location words to describe objects. Therefore, this dataset focuses more on the pure appearance based description. The split in RefCOCO+ follows the same rule used in RefCOCO.

\noindent
(3) \emph{RefCOCOg (Google RefExp)} consists of 85,474 referring expressions for 54,822 objects in 26,711 images from COCO. Different from RefCOCO and RefCOCO+, this dataset is collected using a non-interactive setting and contains much longer sentences. The dataset is randomly partitioned into training, validation (``val'') and testing (``test'') splits.

Since these datasets derive from COCO, we group these images and expressions in each dataset according to 71 entities with high word frequencies in the expressions of COCO. These image groups are shown in Fig. \ref{fig:groups_acc}.

\noindent\textbf{Evaluation Protocols}. For evaluation, we adopt two standard settings in REC \cite{yu2017joint,liu2017referring,yu2018mattnet}. In the first setting, ground-truth bounding boxes are taken as the candidate regions, and a localization is correct if the best-matching region is consistent with the ground-truth. In the second setting, which is more popular now, the candidate regions are extracted by the object detection model, and a localization is correct if the intersection-over-union (IoU) of the best-matching region with the ground-truth bounding box is greater than 0.5. Since REC focuses on visual-textual correspondence and comprehension of cross-modal information, rather than detection performance, we report results under both settings, and conduct analysis and ablation study with the first setting, where accuracy (ACC) is used to evaluate the rate of correct localization. Note that, although the number of the anchor-negative pairs increase, we keep the same number of the training samples and the same batch size of the pairs during model training to ensure the fairness of evaluation.

\begin{figure}[t]
\begin{tabular}{cc}
    \begin{minipage}{0.49\linewidth}
        \leftline{\epsfig{file=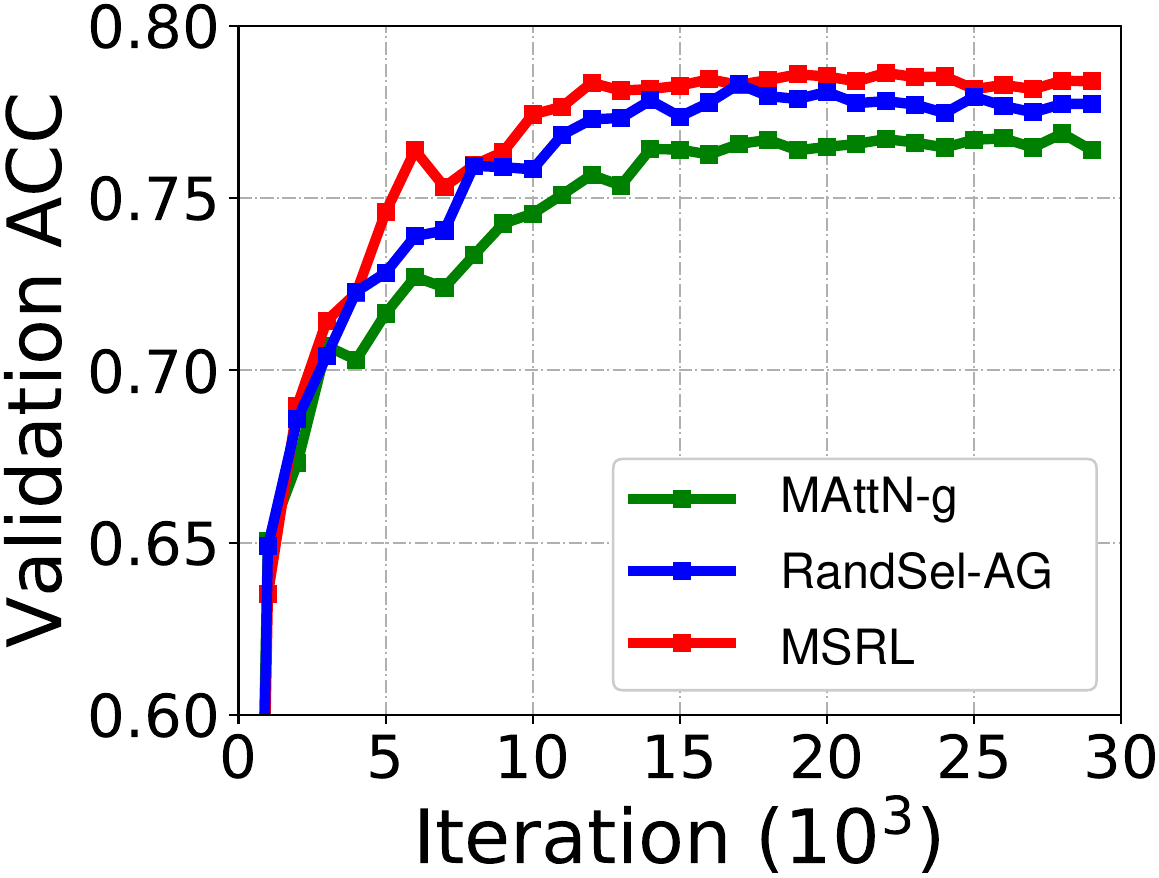, width=1\linewidth}}
    \end{minipage}
    \begin{minipage}{0.49\linewidth}
        \leftline{\epsfig{file=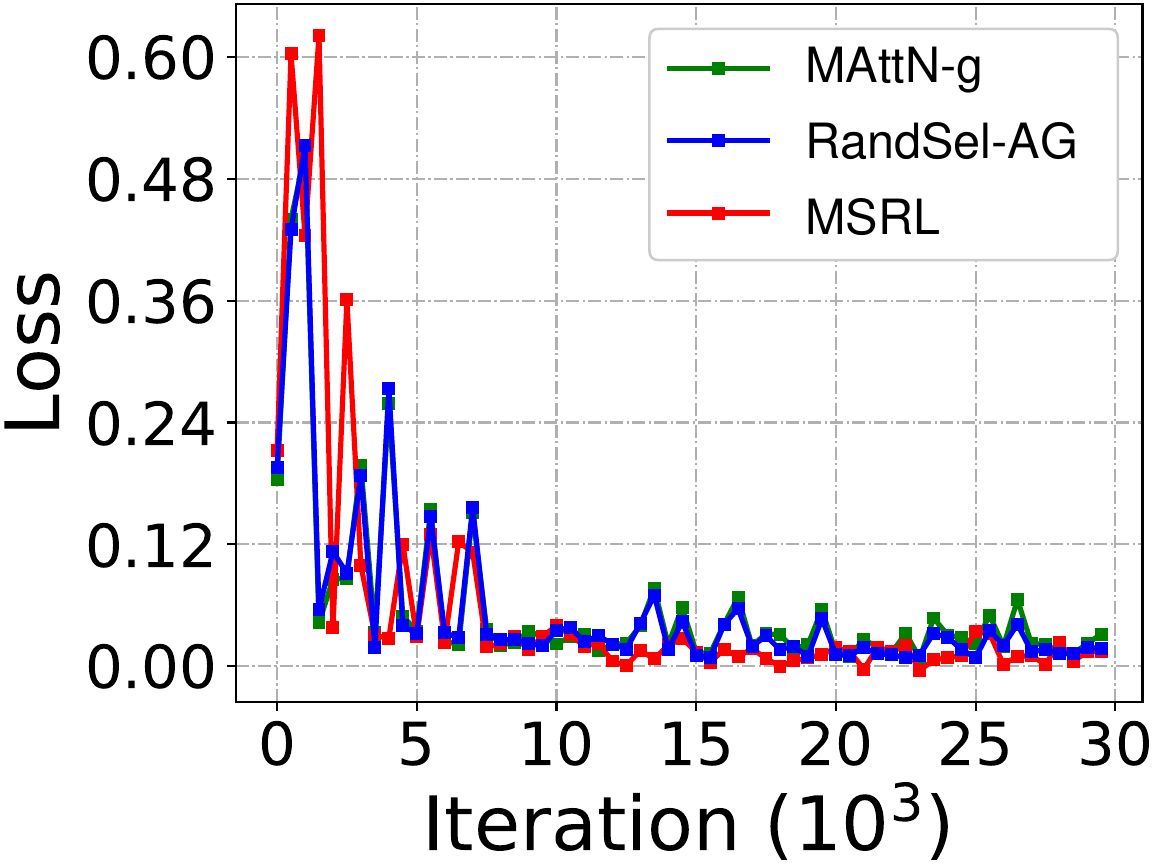, width=1\linewidth}}
    \end{minipage}
\end{tabular}
\caption{Training curves of the proposed MSRL, RandSel-WG, and MAttN-g on the same grouped data. Curves of validation ACC are shown in the right side, and the left is for the curves on training loss. \label{fig:training_curves}}
\end{figure}

\noindent\textbf{Comparison Methods}. We compare the proposed MSRL schema with six baselines: 
(1) MAttN: A joint embedding model that align the region and expression in each image by using multiple modular attention networks \cite{yu2018mattnet} (reproduced by using the official code\footnote{\url{https://github.com/lichengunc/MAttNet}}). 
(2) MAttN-g: An alternative version of MAttN by randomly selecting pairs in our image groups. 
(3) RandSel-WG: An alternative version of MSRL, where the pairs are randomly selected from each group with the same number of pairs. It is set to validate the significance of the differentiated within-group relevance. 
(4) RandSel-AG: An alternative version of MSRL, where the pairs are selected upon their pairwise relevance scores but randomly from different image groups. It is set to validate the significance of the differentiated across-group relevance. 
(5) MSRL-WG: An alternative version of MSRL without the across-group relevance regularization term for the group relevance balance. It is set to validate the significance of group-based relevance learning and the significance of differentiated across-group relevance constraint.
(6) MSRL-AG: An alternative version of MSRL without within-group relevance regularization term. It set to validate the differentiated within-group relevance constraint and its role to further force the anchor and the negative apart.
We also compare the MSRL with the state-of-the-art methods, including the one-stage REC \cite{yang2019fast,yang2020improving,sun2022proposal}, joint embedding based REC \cite{liao2020real}, reinforcement learning based REC \cite{sun2021iterative}, graph-based REC \cite{yang2019dynamic}, and transformer-based REC \cite{du2022visual}.

\subsection{Comparison with the State-of-the-art\label{comparison_SOA}}

There are two experimental settings for the REC evaluation. The first setting assumes that the observer has already known what an object is. Thus, the input region set consists of all the ground truth candidates labeled in the MSCOCO dataset. The model is required to select the anchor object from those ground truth objects. The second setting assumes that the whole process is conducted automatically and proposal-free, where the potential objects are detected from the images in the format of object regions. Such setting is more popular in recent works.

\begin{figure}[t]
\centering
\epsfig{file=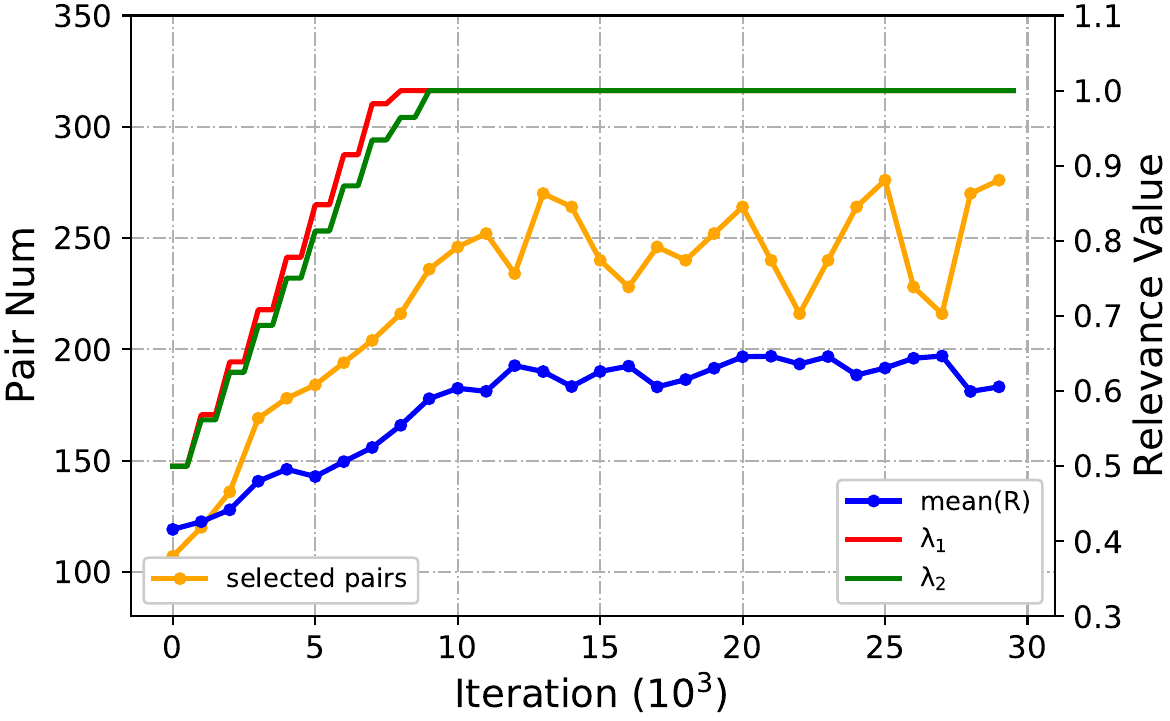, width=1.0\linewidth}
\caption{Changing curves of selected pair number (per batch), average batch relevance ``mean(R)'', parameters $\lambda_1$ and $\lambda_2$ at different iterations. 
``Relevance Value'' is the general estimation for the relevance scores of ``mean(R)'', $\lambda_1$ and $\lambda_2$. \label{fig:changes_figure}}
\end{figure}

We first compare the proposed MSRL model with state-of-the-art methods using MSCOCO's ground-truth bounding boxes as object regions. Results of the comparisons on RefCOCO, RefCOCO+, and RefCOCOg datasets are shown in Tab. \ref{tab:comparison_sota_gt}, which manifests the proposed MSRL achieves the best performance on most evaluation sets. It demonstrates the superiority and effectiveness of MSRL. Specially, MSRL outperforms the baseline MAttN on all metrics, \emph{e.g.}, around $2\%$ gain on the RefCOCOg dataset, which reflects that considering the differentiated within-group and across-group relevances does contribute to the semantic comprehension of regions and expressions. 

Matching the automatically detected objects in REC enables an more practical evaluation in the complete process of comprehension. 
Such evaluation setting is widely followed by the recent methods.
We show the comparison results with the state-of-the-arts using automatically detected objects in Tab. \ref{tab:comparison_sota_det}. Compared to the performances in Tab. \ref{tab:comparison_sota_gt}, the overall performance drops due to the detection errors. Additionally, the proposed MSRL keeps the competitive performances on most evaluation metrics compared to the state-of-the-art methods, \emph{e.g.} the transformer-based VGTR, and other representative methods \emph{e.g.} MAttN, which reflects the robustness of the learning scheme on different data sources and different testing sets. Specially, MSRL outperforms MAttN on three datasets, especially on RefCOCO+ that emphasizes more on object appearance and on RefCOCOg that contains much longer expressions. These manifest that MSRL shows more ability on learning from more complex attributes of subjects.

\begin{table}
	\centering
	\caption{Performance comparisons in the group-based scheme on RefCOCOg. ``Inc.'' denotes the increased performance compared to baseline MAttN. All values are in \%}
	\begin{tabular}{p{1.7cm}<{\raggedright}|p{1.2cm}<{\centering}p{1.2cm}<{\centering}p{1.2cm}<{\centering}p{1.2cm}<{\centering}}
		\hline
		& val & Inc. & test & Inc. \\
		\hline
		MAttN & 77.27 & - & 76.82 & - \\
		MAttN-g & 77.45 & +0.18 & 77.05 & +0.23 \\
		MSRL & 79.15 & +1.88 & 78.42 & +1.60\\
		\hline
	\end{tabular}
	\label{tab:comparison_grouprise} 
\end{table}

\subsection{Ablation Study}

The intuitive ablations of the proposal MSRL are two-fold, \emph{i.e.} on its components and on the data structure. Specially, on one hand, MSRL contains the differential relevance within-group and across-group constraints. It is necessary to conduct the ablation experiments on these two constraints to validate the significance of differential and group-based relevance learning. On the other hand, although the Group-based REC and REC use the same data for training, the former re-organize the data in a group structure, \emph{i.e.} grouped expression-object pairs. Hence, it is necessary to conduct the ablation experiments on the grouped data. 

Therefore, we first compare MSRL with the alternative versions of MSRL without the constraints of the within-group relevance (MSRL-AG) or the across-group relevance (MSRL-WG) in Tab. \ref{tab:comparison_sota_gt}, where MSRL outperforms MSRL-WG and MSRL-AG. This confirms the merits of integrating within-group and across-group relevances in the task of Group-based REC. We also compare MSRL with MSRL-AG and MSRL-WG using automatically detected objects in Tab. \ref{tab:comparison_sota_det}, where MSRL outperforms MSRL-WG and MSRL-AG. It further demonstrates the superiority of integrating within-group and across-group relevances in Group-based REC.

\begin{figure}[t]
\centering
\epsfig{file=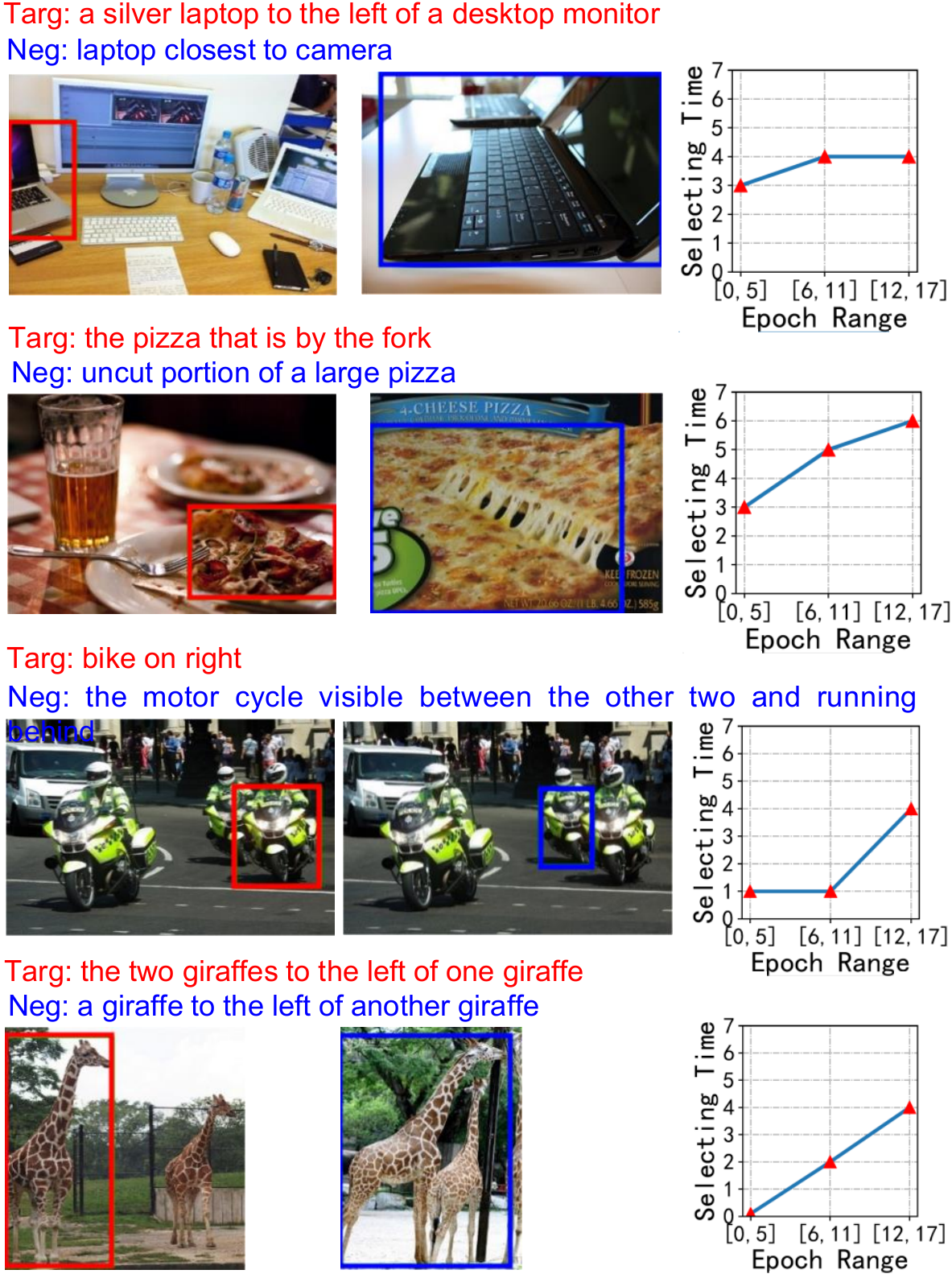, width=1.0\linewidth}
\caption{Examples of the unmatched anchor(Red)-negative(Blue) pairs with the highest (Top) and the lowest (Bottom) diversities. The curves show their selecting times in equidistant epoch ranges. \label{fig:local_pairs}}
\end{figure}

We then let the baseline MAttN use the grouped data and oberse whether the differentiated relevance learning still shows the superior. Tab. \ref{tab:comparison_grouprise} shows that the performance gains of MSRL exceed MAttN-g in the case of group-based scheme compared with MAttN, where MSRL achieves 1.88\% and 1.60\% gains for val and test sets respectively while MAttN-g gets 0.18\% and 0.23\% gains only. 
%
This reflects that our differentiated relevance learning is more superior than the randomly selecting based method. 
Additionally, Tab. \ref{tab:comparison_grouprise} shows that the performance gains of MSRL on both val and test sets are much larger than double the performance gains of MAttN-g, which indicates the role of differentiated relevance leanring is much more significant than data grouping. 
The reason may be that, even though the number of the negatives increase significantly from a single image to a image group for each anchor, the effective anchor-negative pairs, \emph{e.g.} the one in the image of the anchor, are simultaneously diluted relative to the total number. The thrust thus becomes how to select and use the pairs properly with the training going on.

Besides the above two intuitive ablations, deeper ablations are carried out in the next two sections to analyze the roles of the differentiated within-group relevance and the differentiated across-group relevance respectively.

\subsection{Analysis on Within-group Relevance\label{subsec:ablation_pairwiseDiversity}}

\begin{figure}
\centering
\epsfig{file=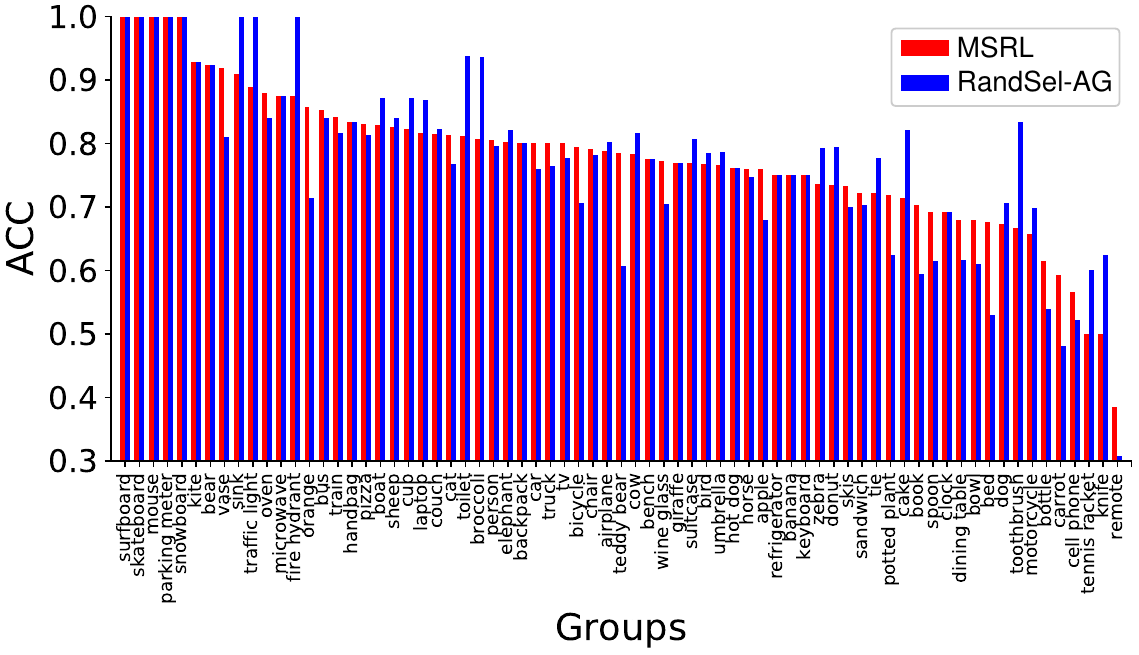, width=1.0\linewidth}
\caption{Group-wise comparisons on ACC between the methods with and without considering the across-group relevance (MSRL vs. RandSel-AG). \label{fig:groups_acc}}
\end{figure}

We analyze the within-group relevance from three aspects, \emph{i.e.}, the training curves, the change of data during the selection, and the different examples with different relevance scores.
Specially, we first present the training curves of different methods over the entire iterations in Fig. \ref{fig:training_curves}. These curves show that the performance of MSRL increases faster than RandSel-WG at the beginning, which manifests the differentiated within-group relevance learning schema selects and uses the anchor-negative pairs effectively. 
Additionally, MSRL reaches the best value among the three methods on the validation accuracy after around 2,500 iterations, which verifies the superiority of the relevance-guided data selection and our learning scheme. Although the loss curve of MSRL fluctuates obviously in the early stage, it converges faster and more smoothly after around 7,500 iterations, 
which indicates that the adaptive learning on the pairs from low relevance to high relevance does contribute to the model optimization. The smooth and low loss curve of MSRL also reveals that MSRL can effectively enlarge the relevance gap and keep the margin stably between the anchor-positive and the anchor-negative pairs by distinguishing the anchors/positives and the negatives.

Secondly, since within-group relevance has the essential role in the data selection, we present the changing curves of the number of the selected unmatched pairs, the average batch relevance, and the parameters $\lambda_1$ and $\lambda_2$ at different iterations in Fig. \ref{fig:changes_figure}. During the model training, more and more unmatched pairs are selected with the growth of $\lambda_1$ and $\lambda_2$, which shows the stability and effectiveness of MSRL on controlling the pace of pair selection the pairs based on within-group relevance. Moreover, the average batch relevance grows gradually in the early stage of training and turns to level off, while the selected pair number grows observably in the early training stage and turns to sharp fluctuation afterwards. These confirm that the pairs with low relevances are selected in the early stage of training and the pairs with high relevances are selected gradually, during which the number of the selected pairs doesn't always grow when $\lambda_1$ and $\lambda_2$ reach the maximums. Besides, Fig. \ref{fig:training_curves} shows that MSRL outperforms the other two at the beginning, specifically at around iteration $5\times10^3$, while few pairs are selected (see the cyan curve in Fig. \ref{fig:changes_figure}), which reflects that the (low) relevance of the data, rather than its scale or its variety, affects the performance. The pairs with higher relevances are gradually selected, which can be seen as an adaptive sampling process to overcome the overfitting. It further reveals the robustness and the adaptiveness of the proposed MSRL model on selecting complex pairs.

\begin{table}[t]
\centering
\caption{The effect of the across-group relevance regularization (AG-R) on pair selection over different groups. ``M'' denotes the mean numbers of the selected pairs among groups in different iteration ranges (``Iter Range'', $\times10e^3$). Here, ``CV'' (Coefficient of Variation \cite{abdi2010coefficient}) denotes the ratio (\%) of the standard deviation of pair numbers over different groups to their mean number}
\begin{tabular}{p{0.8cm}<{\centering}p{0.22cm}<{\centering}|p{0.6cm}<{\centering}p{0.6cm}<{\centering}p{0.7cm}<{\centering}p{0.7cm}<{\centering}p{0.7cm}<{\centering}p{0.7cm}<{\centering}}\hline
\multicolumn{2}{c|}{Iter Range} & (0,5] & (5,10] & (10,15] & (15,20] & (20,25] & (25,30] \\
\hline
w/o & M   & 12598 &  12835 & 12853 & 12857 & 12852 & 12854 \\ \cline{2-8}
AG-R & CV & 88.64 &  88.57 &  88.86  &  88.46  &  88.74  &  88.49 \\ \hline
w & M   & 13120 &  13377 & 13391 & 13393 & 13397 & 13385 \\ \cline{2-8}
AG-R & CV & 86.15  &  84.73  &  83.60  &  83.34  &  83.48  &  85.29 \\ \hline
\end{tabular}
\label{tab:group_pair_num_mean_CV}
\end{table}

Finally, we show the examples of the unmatched pairs with the lowest and highest relevance scores in Fig. \ref{fig:local_pairs}. Comparing the visual and the textual contents, the relevances of the pairs between the top two and the bottom two rows vary greatly. For example, the two \emph{pizzas} (in the second row) appear less relevant compared to the \emph{giraffe} pairs (in the bottom row). It verifies the ability of MSRL on capturing the relevance. Moreover, as the curves show, the pairs with low relevance scores are selected more often in the beginning, while the pairs with high relevance scores are selected gradually up to the similar times as the formers, which further validates the effectiveness of our low-to-high relevance based sample selection scheme.

\subsection{Analysis on Across-group Relevance\label{subsec:ablation_groupwiseDiversity}}

\begin{figure}[t]
    \begin{minipage}{0.49\linewidth}
        \leftline{\epsfig{file=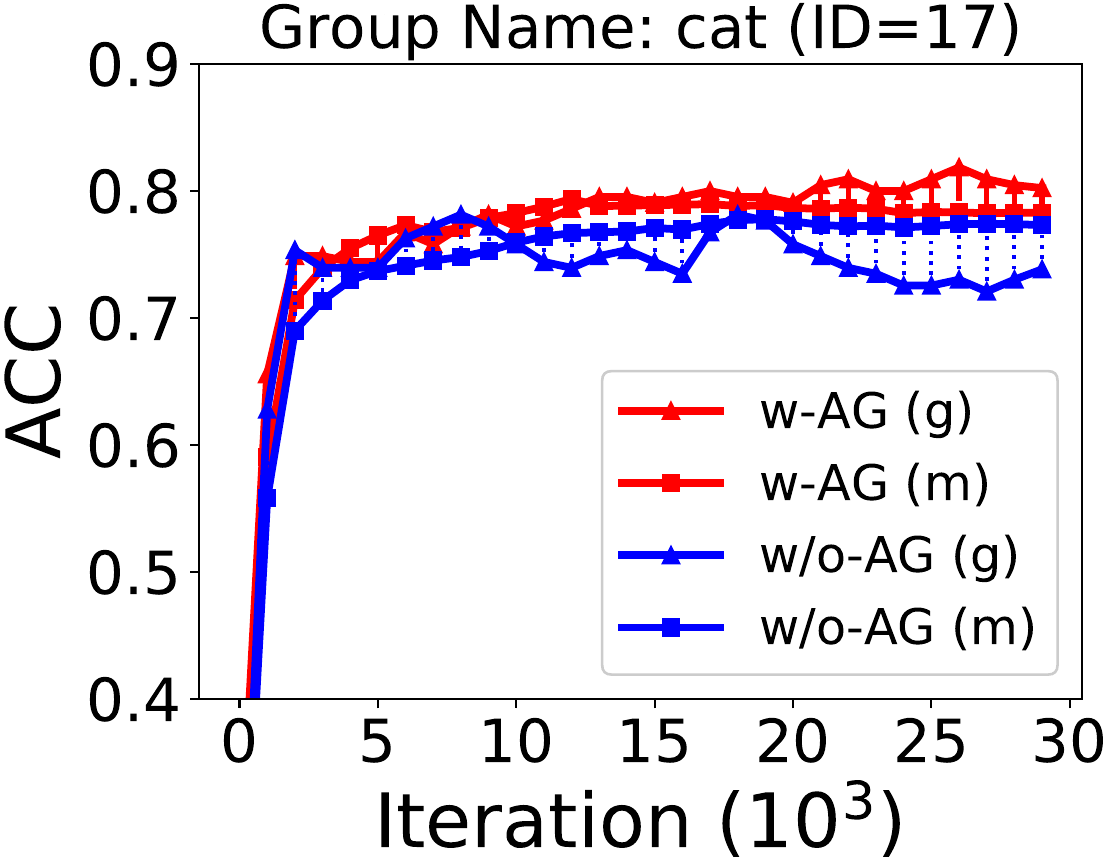, width=1\linewidth}}
    \end{minipage}
    \hfill
    \begin{minipage}{0.49\linewidth}
        \leftline{\epsfig{file=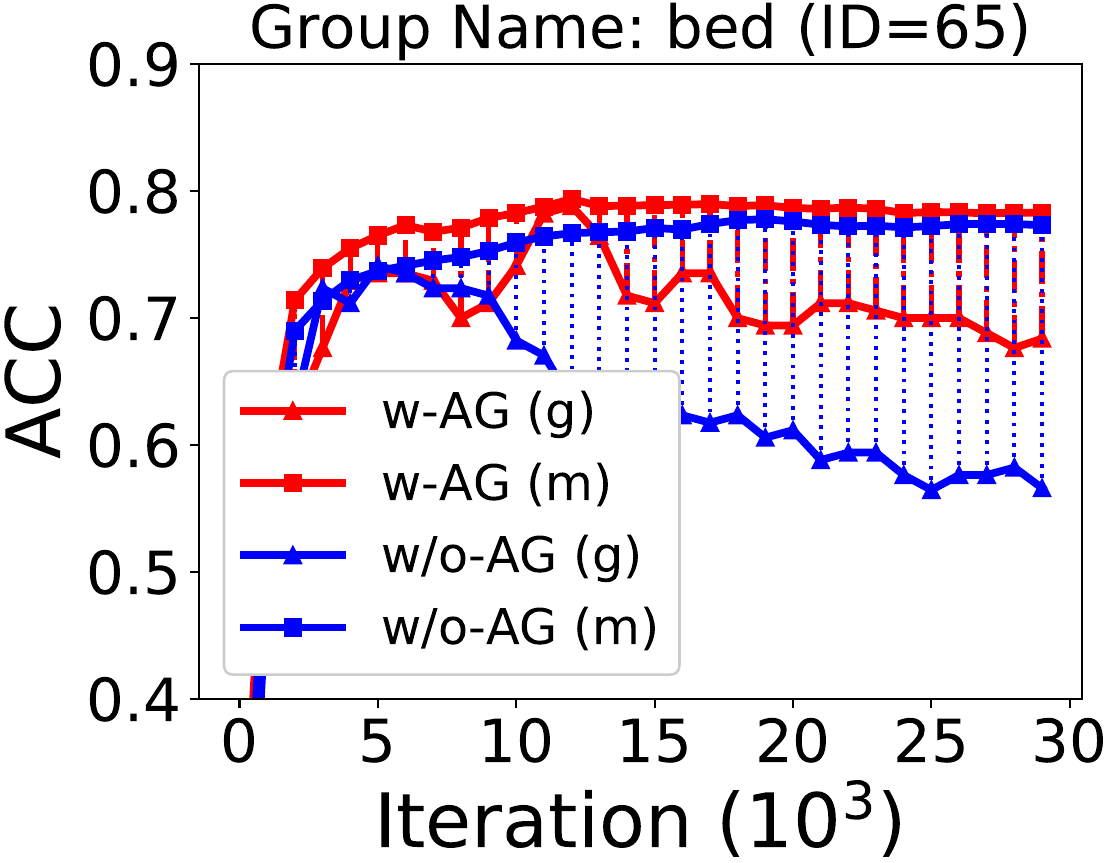, width=1\linewidth}}
    \end{minipage}
    \vfill
    \begin{minipage}{0.49\linewidth}
        \leftline{\epsfig{file=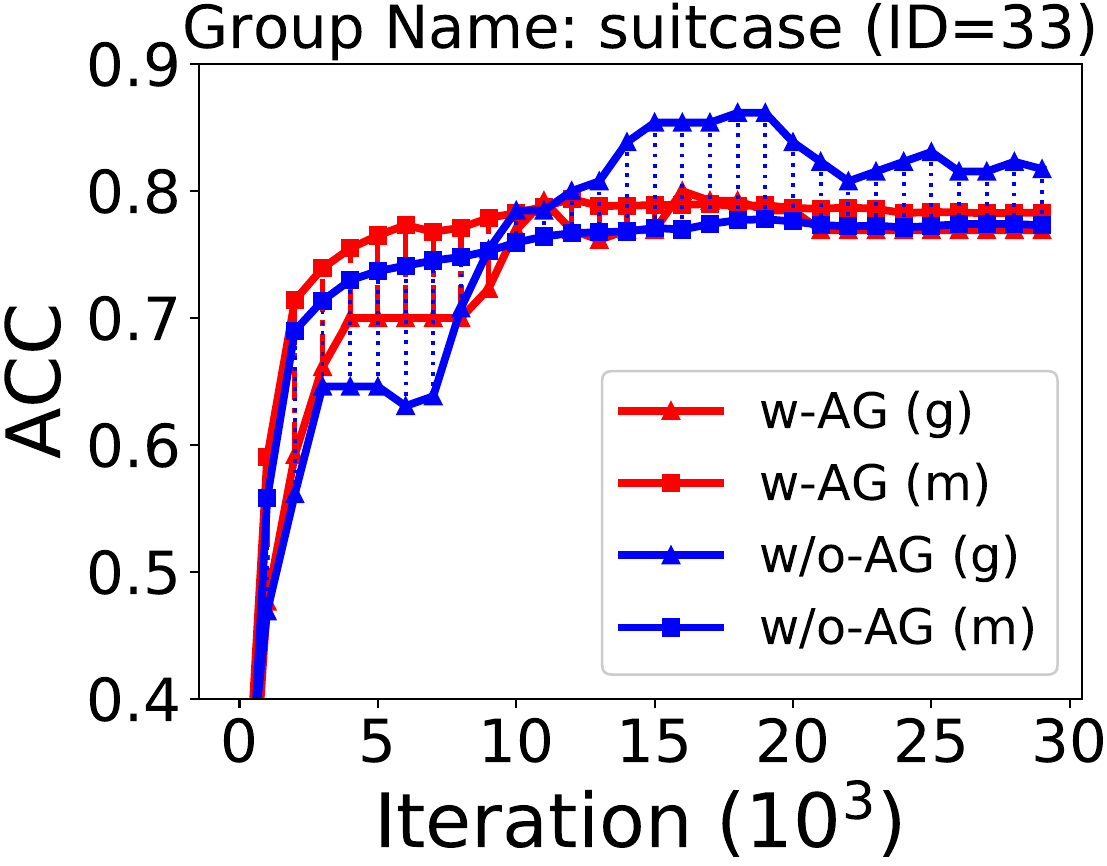, width=1\linewidth}}
    \end{minipage}
    \hfill
    \begin{minipage}{0.49\linewidth}
        \leftline{\epsfig{file=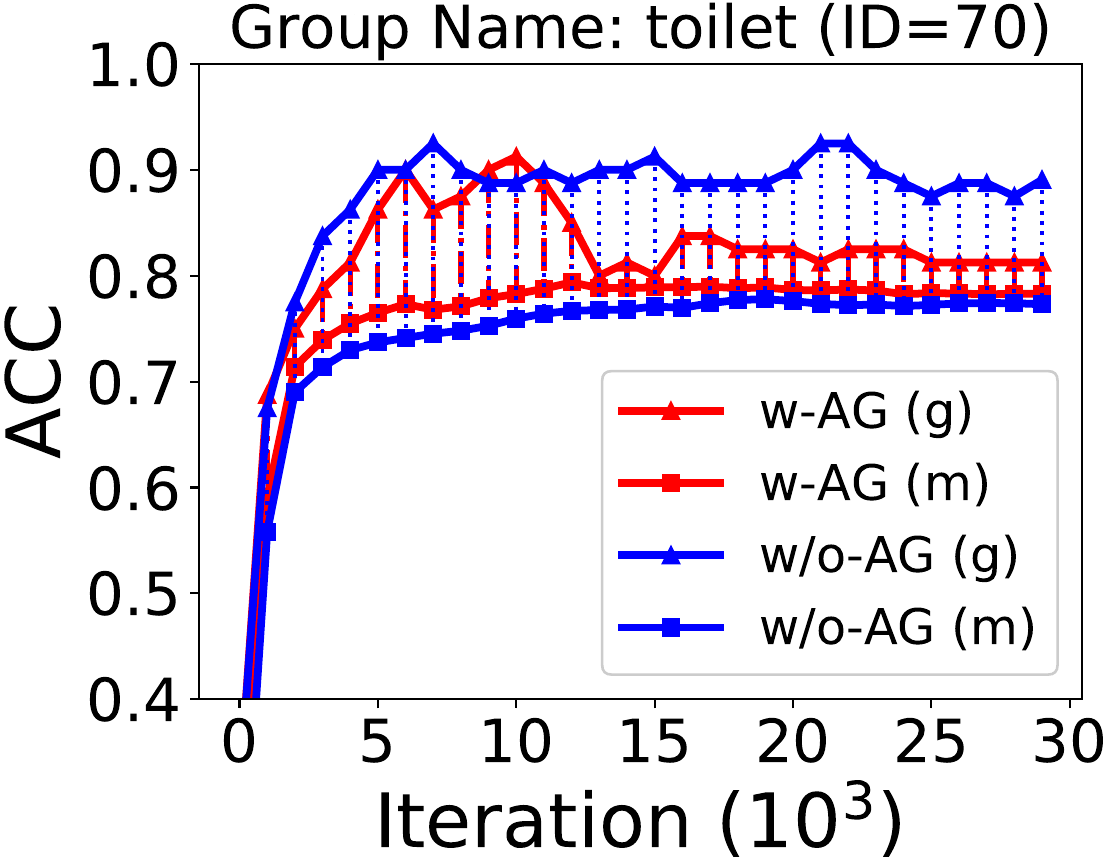, width=1\linewidth}}
    \end{minipage}
\caption{Group-specific validation curves of average ACCs of some example groups at different iterations, which are marked with triangle markers and ``(g)''. Curves with square markers are the average ACC of all the groups (``(m)''). The red and blue colors denote MSRL (``w-AG'') and RandSel-AG (``w/o-AG'') respectively. 
The top two examples are the cases when MSRL has higher performance on the group ACC while the bottom two are converse. \label{fig:groups_curve}
}
\end{figure}

\begin{table}[t]
\centering
\caption{Comparisons on the average ACC (AVE) and the corresponding standard deviation (STD) over different groups at different training iterations. All values are in \%}
\begin{tabular}{p{0.6cm}<{\raggedright}p{1.7cm}<{\centering}|p{0.9cm}<{\centering}p{0.9cm}<{\centering}p{0.9cm}<{\centering}p{0.9cm}<{\centering}}\hline
\multicolumn{2}{c|}{\multirow{2}{*}{}}
& Iter & Iter & Iter & Iter \\
\multicolumn{2}{c|}{} & (0.5K) & (1K) & (5K) & (10K) \\
\hline
\multirow{2}*{AVE}
& RandSel-AG   & 60.07 & 65.75 & 74.05 & 75.94 \\ \cline{2-6}
& MSRL & 62.45 & 66.56 & 75.30 & 77.26 \\ \hline
\multirow{2}*{STD}
& RandSel-AG   & 17.81 & 15.30 & 13.01 & 11.88  \\ \cline{2-6}
& MSRL & 17.26 & 14.80 & 12.67 & 10.78  \\ \hline
\end{tabular}
\label{tab:group_ave_std}
\end{table}

We conduct four subdivided experiments to analyze the role of across-group relevance in the proposed MSRL schema. Specially, we first compare the accuracies of different groups with and without the constraint of the across-group relevance in Fig. \ref{fig:groups_acc}, from which we find that the performances of different groups are more balanced after considering the constraint of the across-group relevance during model training. It reflects the effectiveness of the constraint to reduce the bias of specific categories. Additionally, the red redundant area is apparently larger than the blue one, which indicates the superiority of the proposed MSRL on the performances among most groups.

\begin{figure}[t]
\centering
\epsfig{file=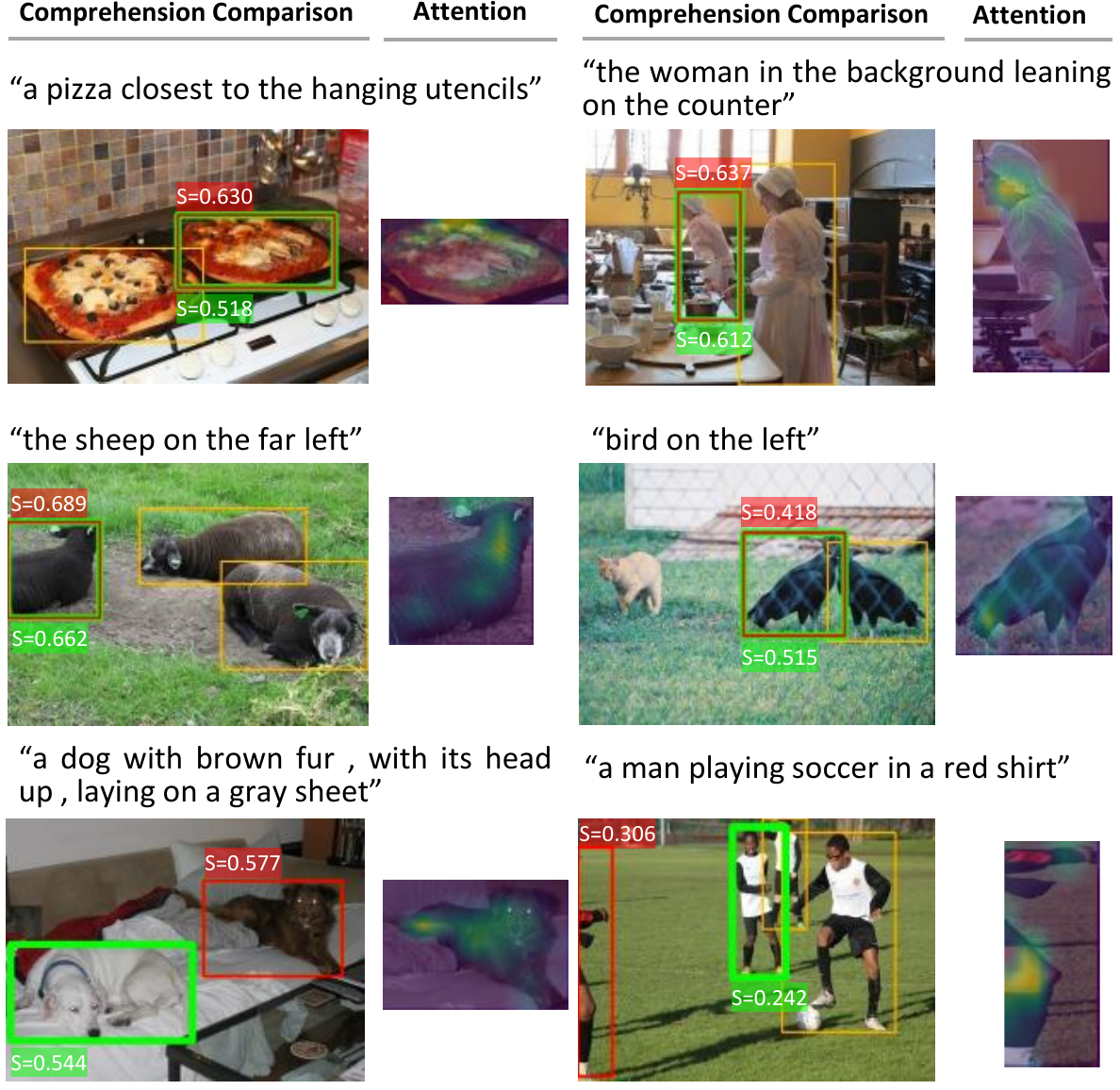, width=1.0\linewidth}
\caption{Relevance visualization before and after our multi-group self-pace relevance learning. The red and green boxes with matching scores are the localization results of the proposed MSRL and MAttN, respectively. The orange boxes denotes the region candidates. The heat maps show our visual attention scores guided by the expressions. \label{fig:visualization_comparison}}
\end{figure}

Then, we analyze the internal effect of the across-group relevance regularization $\|\bm{\mathcal{U}}\|_{F,1}$ in Tab. \ref{tab:group_pair_num_mean_CV}. Since across-group relevance regularization decides the priority balance of pairs, we can analyze it on the numbers of the selected pairs for different groups. Tab. \ref{tab:group_pair_num_mean_CV} shows the mean number of the selected pairs among groups in different iteration ranges, as well as the ratio of the standard deviation of the pair numbers (over different groups) to their mean number. We find that the mean numbers with across-group relevance regularization in different ranges are larger than the ones without the regularization, which reflects the counter-sparsity effect of the regularization on pair selection over groups. Besides, the ``CV'' with across-group relevance regularization in each range is lower than the ones without the regularization, which manifests the ability of the regularization on balancing the priorities of different groups. Moreover, with the iteration increasing, the ``CV'' with the regularization tends to flatten. This indicates the controllability of the across-group relevance balancing in the overall schema.

Next, we further analyze the average validation ACC of different groups at different iterations in Fig. \ref{fig:groups_curve}. Whether RandSel-AG outperforms MSRL or not on the ACC of each group, the group ACC curves of MSRL tend to be close to its average ACC curves, which reflects the effect of the across-group relevance constraint on balancing the multi-group training. Additionally, from the average ACC curves, we find that MSRL is enhanced by the across-group relevance constraint, which surpasses RandSel-AG during almost the entire training period. This reflects the overall effect of the across-group relevance.

Finally, we compare the proposed MSRL with RandSel-AG on the average ACC (AVE) and the corresponding standard deviation (STD) over different groups in Tab. \ref{tab:group_ave_std}. We compute the AVEs and STDs at the 0.5K-th, 1K-th, 5K-th, and 10K-th iterations, respectively. Clearly, the AVEs of the proposed MSRL are higher than the ones of RandSel-AG, but with much smaller STDs. It demonstrates that the across-group relevance constraint not only boosts the overall learning performance, but also reduces the bias of certain categories.

\subsection{Relevance Visualization}

To observe the expression-region matching performance after our differentiated relevance learning, we show the visualized examples Fig. \ref{fig:visualization_comparison} with the expression-region matching scores respectively generated by the proposed MSRL and MAttN. In the top two rows, both two methods can locate the correct regions. However, with respect to some complex pairs in the bottom row, \emph{e.g.}, long expression (the right-bottom case) and missing parts (the left-bottom case), MSRL achieves better comprehension and shows the robustness. This may be because MSRL conducts the coarse-to-fine adaptive learning to distinguish more various anchor-negative pairs.

\section{Conclusion\label{sec:conclusion}}

In this paper, we propose a new problem of referring expression comprehension, \emph{i.e.} Group-based REC, where the negatives of each anchor is augmented across images according to the fact that there could be multiple attributes in each anchor expression/object and multiple attribute values for each attribute. 
%
To learning from the abundant negatives that have the differentiated relevances to their anchors,
%
we propose a multi-group self-paced relevance learning method (termed MSRL) based on the differentiated within-group relevance and the differentiated across-group relevance, 
In MSRL, the within-group expression-object pairs are adaptively assigned with different priorities according to their cross-modal relevance scores, and the bias of the group priority is balanced via an across-group relevance constraint. The relevance scores, the pairs, and the model parameters are alternately updated in the end-to-end adaptive model training. We have demonstrated that the proposed MSRL achieves superior performance over the state-of-the-art REC methods on three standard benchmarks (RefCOCO, RefCOCO+, and RefCOCOg). We have also conducted extensive ablative experiments to further demonstrate the effectiveness of MSRL on learning differentiated within-group relevance and differentiated across-group relevance.

In the future, it is fascinating to further explore Group-based REC on the different aspects: 
(1) In practice, some image groups may be missing in the training set, while some attributes or some attribute values may also be missing among the anchor-negative pairs of the training set, which suppresses the semantic learning.
It would be inspirational to advance the proposed adaptive learning strategy to model the group-wise correlation to convey the representations of the attributes. 
(2) Some computer vision tasks like Person Re-identification \cite{ye2021deep,gu2022clothes} can also be transformed into group-based problems, where the semantics of people in a same group are similar and the relevances of the anchor-negative pairs are differentiated. 

\section*{Acknowledgment}

This work was supported by Fujian Province Science and Technology plan project (No. 2020Y4010 and No. 2022Y4012).

\printcredits

\bibliographystyle{elsarticle-num}
\bibliography{cas-refs}

\end{document}